\documentclass[journal, twoside]{IEEEtran}
\usepackage[flushleft]{threeparttable}

\usepackage{array}
\usepackage{booktabs}
\usepackage{multicol}
\usepackage{epstopdf}

\usepackage{arydshln}
\usepackage{color}
\usepackage{fancyhdr}
\usepackage{amsthm}
\usepackage{algorithmic}
\usepackage{algorithm2e}

\usepackage{lipsum}
\usepackage{multirow}
\usepackage{amsfonts}
\usepackage{amsmath}
\usepackage{siunitx}

\usepackage[nocompress]{cite}

\ifCLASSINFOpdf
\usepackage[pdftex]{graphicx}
\graphicspath{{Figures/}}
\DeclareGraphicsExtensions{.pdf,.jpeg,.png}
\else
\usepackage[dvips]{graphicx}
\graphicspath{{Figures/}}
\fi
\usepackage{pgfplots}
\usepackage{url}

\begin{document}
%\pgfplotsset{compat=1.15}
\title{On Improving Hotspot Detection Through \\ Synthetic Pattern-Based Database Enhancement}

\author{Gaurav~Rajavendra~Reddy,~\IEEEmembership{Student Member,~IEEE},
        Constantinos~Xanthopoulos,~\IEEEmembership{Student Member,~IEEE},
        Yiorgos~Makris,~\IEEEmembership{Senior Member,~IEEE}%
\thanks{G. R. Reddy, C. Xanthopoulos, and Y. Makris are with the Department of Electrical and Computer Engineering, The University of Texas at Dallas, Richardson, TX 75080 USA (e-mail: gaurav.reddy@utdallas.edu; constantinos.xanthopoulos@utdallas.edu; yiorgos.makris@utdallas.edu).}
}

% The paper headers
\markboth{}{REDDY \MakeLowercase{\textit{et al.}}: ON IMPROVING HOTSPOT DETECTION THROUGH SYNTHETIC PATTERN-BASED DATABASE ENHANCEMENT}

% make the title area
\maketitle

\begin{abstract}
Continuous technology scaling and the introduction of advanced technology nodes in Integrated Circuit (IC) fabrication is constantly exposing new manufacturability issues. One such issue, stemming from complex interaction between design and process, is the problem of design hotspots. Such hotspots are known to vary from design to design and, ideally, should be predicted early and corrected in the design stage itself, as opposed to relying on the foundry to develop process fixes for every hotspot, which would be intractable. In the past, various efforts have been made to address this issue by using a known database of hotspots as the source of information. The majority of these efforts use either Machine Learning (ML) or Pattern Matching (PM) techniques to identify and predict hotspots in new incoming designs. However, almost all of them suffer from high false-alarm rates, mainly because they are oblivious to the root causes of hotspots. In this work, we seek to address this limitation by using a novel database enhancement approach through synthetic pattern generation based on carefully crafted Design of Experiments (DOEs). Effectiveness of the proposed method against the state-of-the-art is evaluated on a 45nm process using industry-standard tools and designs.
\end{abstract}

% Note that keywords are not normally used for peerreview papers.
\begin{IEEEkeywords}
Lithographic Hotspot Detection, Synthetic Pattern Generation, Design For Manufacturability, Database Enhancement, Machine Learning.
\end{IEEEkeywords}

\IEEEpeerreviewmaketitle

\section{Introduction} \label{sec:intro}
\begin{figure}[t!]
	\centering
	\includegraphics[clip=true, trim=0.06in 0.2in 0.04in 0.2in, width=0.50\textwidth]{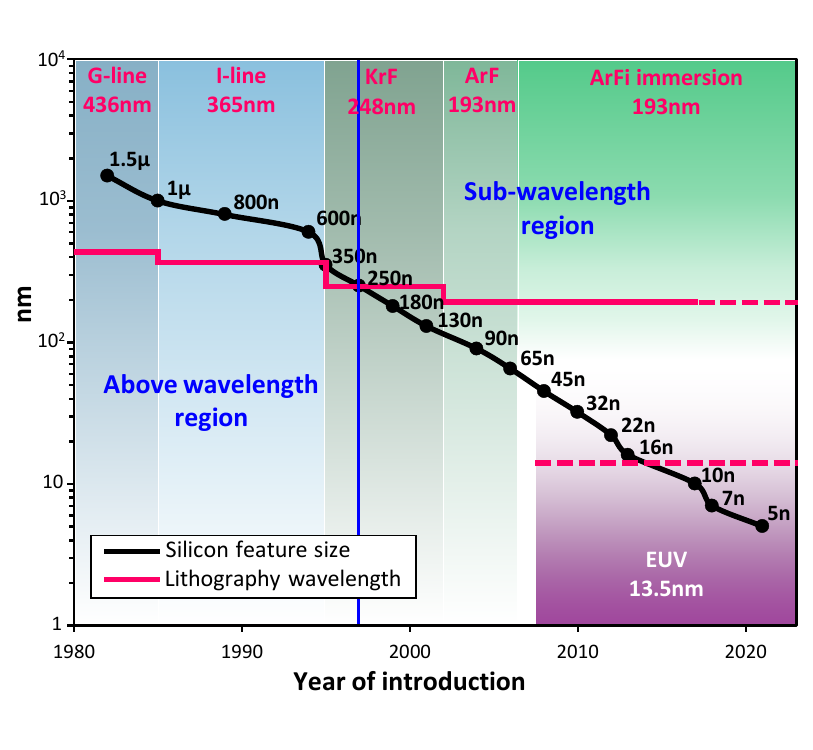}
	\vspace{-0.075in}
	\caption{Changes in lithography with silicon feature sizes (Adapted from \cite{LithoTrendLeuven2018})}
	\label{fig:litho_trend}
	\vspace{-0.15in}
\end{figure}

Continued technology scaling and the introduction of every advanced technology node in Integrated Circuit (IC) fabrication brings in new challenges for foundries. Among them, lithography is a major obstacle during new technology development. As shown in Figure \ref{fig:litho_trend}, in early technology nodes, the wavelength of light used in lithography was much smaller than the features being printed. In the latest nodes, however, this is no longer the case and lithography has become extremely challenging due to complex interactions between designs and sophisticated unit processes. To mitigate some of the lithography-related issues and ensure reliable manufacturing, various Resolution Enhancement Techniques (RETs) such as Optical Proximity Correction (OPC), Multi-patterning, Phase-shifted masks, etc., are used. Despite employing RETs, certain areas in the design (layout), which pass Design Rule Checks (DRCs) and comply with Design For Manufacturability Guidelines (DFMGs), show abnormal and unexplained variation, causing parametric or hard defects. Such areas are termed as `Hotspots' (popularly known as `Lithographic hotspots' or `Design weak-points'). The cause of hotspots is mostly attributed to their neighborhood (i.e., a set of polygons surrounding the hotspot area) which causes complex interactions of light during the lithography process. Since hotspots vary from design to design, identifying their root causes and finding a fix for all such hotspots through process changes is extremely difficult, time consuming and expensive. Thus, in most cases, foundries create a database of known hotspots and restrict their presence in incoming customer designs. A hotspot database is usually populated through Failure Analysis (FA), inline inspections, lithographic simulations using well-calibrated lithographic models, etc. \cite{DRCplus}. If a design pattern turns out to be a hotspot in later stages of fabrication, especially after mask production, it may result in large financial losses to the foundry. Hence, there is a great incentive to identify hotspots early and correct them in the design stage itself.

%In spite of employing RETs, complex designs from different designers/EDA-tools give rise to various manufacturability issues which are often not foreseen by the foundry. One of such issues is Design Hotspots/weak-points or popularly known as lithographic hotspots. Certain areas in the design show abnormal and unexplained variation despite passing Design Rule Checks (DRCs) and complying with Design For Manufacturability Guidelines (DFMGs). Such areas are termed as hotspots. The cause of these hotspots is mostly attributed to their neighborhood (a set of polygons surrounding the hotspot area) which causes complex interactions of light during the lithography process. Since these hotspots vary from design to design, the process of identifying the root cause of such abnormal effects of light and finding a fix for all such hotspots through changes in the process is extremely difficult, time consuming and expensive. Thus, in most cases, foundries create a database of such known hotspots and restrict their presence in incoming customer designs. Foundries usually populate such a database through either Failure Analysis (FA) data, inline inspections, through lithographic simulations using well calibrated lithographic models \cite{DRCplus} etc. If a design pattern turns out to be a hotspot in later stages of fabrication, especially, after mask making, it may result in huge financial losses to the foundry. Hence, there is a great need to identify these problematic patterns (hotspots) early, and correct them in the design stages itself.

\begin{figure*}[t!]
	\centering
	\vspace{-0.125in}
	\includegraphics[clip=true, trim=0.2in 0.2in 0.2in 0.2in,width=0.95\textwidth]{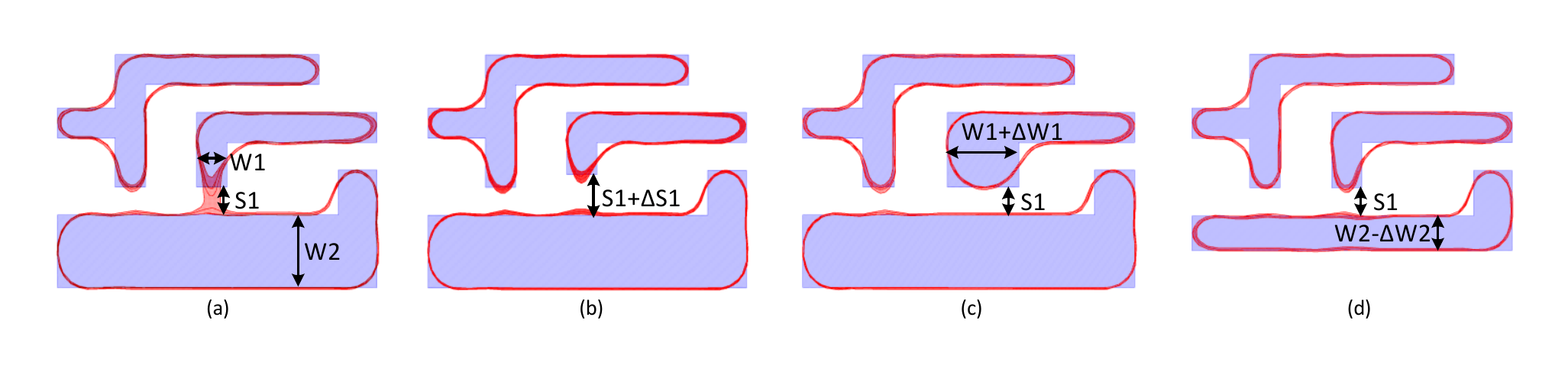}
	\vspace{-0.20in}
	\caption{(a) A hotspot pattern, (b-d) variants of pattern (a) which are non-hotspots}
	\label{fig:case_study}
	\vspace{-0.095in}
\end{figure*}
Many researchers have suggested pattern matching and machine learning-based techniques to identify and predict hotspots in new incoming designs. Unlike previous works \cite{DPAN_ASPDAC,DPAN_TCAD11}, where the focus has been on using increasingly more powerful machine learning tools, we take a novel approach to improving hotspot detection by increasing the information-theoretic content of the training data that these methods use. We call this process `Database enhancement' and it involves two procedures, namely \textit{synthetic pattern generation} and \textit{design of experiments}. Combined, these procedures enable a machine learning entity to effectively learn the `root cause features' of hotspots. These procedures are also `method agnostic', as they can be used with any of the previously proposed hotspot detection methods to improve their performance.
%and those methods would potentially perform better than before.

The rest of the paper is organized as follows. The State-Of-The-Art (SOTA) and its limitations are explained in detail in Section \ref{sec:SOTA}. The proposed methodology is presented in Section \ref{sec:flow}. Experimental results and a discussion are presented in Sections \ref{sec:expresults} and \ref{sec:discuss}, respectively. Conclusions are drawn in Section \ref{sec:conclusion}.

\section{The State-Of-The-Art And Its Limitations} \label{sec:SOTA}

In this section, we briefly review the state of the art in hotspot detection and we pinpoint its key limitation, namely the high rate of false alarms.

\subsection{Hotspot Detection Literature Review} \label{subsec:review}

Hotspot detection has been a topic of high interest in the past decade. Authors of \cite{fuzzy_PM,PMATCH2006} have used Pattern Matching (PM) techniques, wherein a new design is compared to a database of previously seen hotspots and potential hotspot areas in the design are flagged. While these techniques are helpful in quickly analyzing large layouts and identifying known hotspots, they also cause large amounts of false alarms. To address this issue, Machine Learning (ML-) based methods were proposed. These methods essentially `learn' (are trained) from a known database and use the trained model to make a prediction on new patterns. In the past decade, several variants of ML-based hotspot detection methods were introduced and most of them have shown slight improvements over their predecessors in terms of hotspot hit rates and false alarm rates. Such methods focused on using increasingly more powerful ML-based algorithms, wherein the usage of Support Vector Machines (SVMs) \cite{DPAN_TCAD11}, Artificial Neural Networks (ANNs) \cite{DPAN_ICICDT}, multiple/meta classifiers \cite{DPAN_ASPDAC}, Adaboost classifiers \cite{matsunawa2015new}, etc., was proposed. Other sophisticated methods including online learning \cite{BEIYU_ICCAD_16}, wire distance-based feature extraction \cite{kataoka2018novel}, litho-aware learning \cite{park2018litho}, hybrid PM-ML solutions \cite{MADKOUR_16}, etc., have also shown improved results. More recently, deep learning-based methods were proposed \cite{borisov2018lithography}\cite{yang2017DAC}\cite{yang2018TCAD}{\color{black}\cite{semisupervise2019}\cite{lithoroc2019}}. Authors of \cite{yang2017DAC} proposed the use of \textit{feature tensors}, which retain spatial relationships between features, along with biased learning and batch biased learning \cite{yang2018TCAD}. In \cite{yang2017imbalance}, imbalance-aware deep learning has been proposed to address the issue of disproportionate cardinality of hotspots and non-hotspots in the training datasets. Most of these methods, however, still suffer from high false-alarm rates when exposed to Hard-To-Classify (HTC) patterns. This is mainly because these techniques are oblivious to the root causes of hotspots and ignore the fine $nm$-level differences between similar-looking patterns, which play a significant role in making a pattern a hotspot or a non-hotspot. In contrast, the DataBase (DB) enhancement approach proposed herein seeks to specifically address this limitation.

\subsection{False-Alarms In The State-Of-The-Art}\label{subsec:falarms}
The state-of-the-art machine learning-based hotspot detection techniques suffer from high false-alarm rates\cite{fuzzy_PM,DPAN_ASPDAC,DPAN_TCAD11,BEIYU_ICCAD_16}. The source of these false-alarms is illustrated using the following example. Figure \ref{fig:case_study} shows four patterns with their contours (Process Variability (PV) bands) obtained from lithography simulations. Among them, pattern (a) is a hotspot due to a short between two of its polygons. Patterns (b-d) are very similar to pattern (a), but their subtle differences from pattern (a) makes them non-hotspots.

\textbf{Case 1} - Let us assume that an ML-based classifier is being trained to detect hotspots and that, among the patterns shown in Figure \ref{fig:case_study}, only pattern (a) is part of its training dataset. During testing, if pattern (b) is presented to the classifier, it tends to classify it as a hotspot due to its close similarity to pattern (a). But, in reality, it is not a hotspot due to the increased space $S1 + \Delta S1$. The classifier made this error because it had failed to recognize $S1$ as a root cause feature of this pattern. 

\textbf{Case 2} - Let us assume that the classifier's training dataset includes both patterns (a) and (b). In this case, the classifier easily recognizes that the constrained space $S1$ makes this pattern a hotspot and a relaxed space $S1 + \Delta S1$ would make it a non-hotspot. Then, if pattern (c), which is very similar to patterns (a) \& (b) (also having a constrained space $S1$), is presented to the classifier, the classifier tends to call it a hotspot. But in reality, it is not a hotspot because of the increased width $W1 + \Delta W1$.  Here, the classifier predicted incorrectly because, during training, it had only recognized $S1$ as a root cause feature, but not $W1$. Similarly, the feature $W2$, which is crucial for determining that pattern (d) is not a hotspot, is also a root cause feature.

From the above example, it becomes evident that, unless otherwise trained with many variants of a known hotspot, the ML entity assumes that all polygons in a pattern contribute equally towards making it a hotspot and fails to learn the root cause features. Without such learning, it remains oblivious to the subtle variations in similar-looking patterns and tends to misclassify them, creating large amounts of false-alarms. Hence, enhancing the database with sufficient variants of known hotspots becomes imperative towards empowering an ML entity to learn effectively. 

\begin{figure*}[t!]
	\centering
	\vspace{-0.15in}
	\includegraphics[clip=true, trim=0.0in 0.0in 0.0in 0.0in,width=\textwidth, scale=0.98]{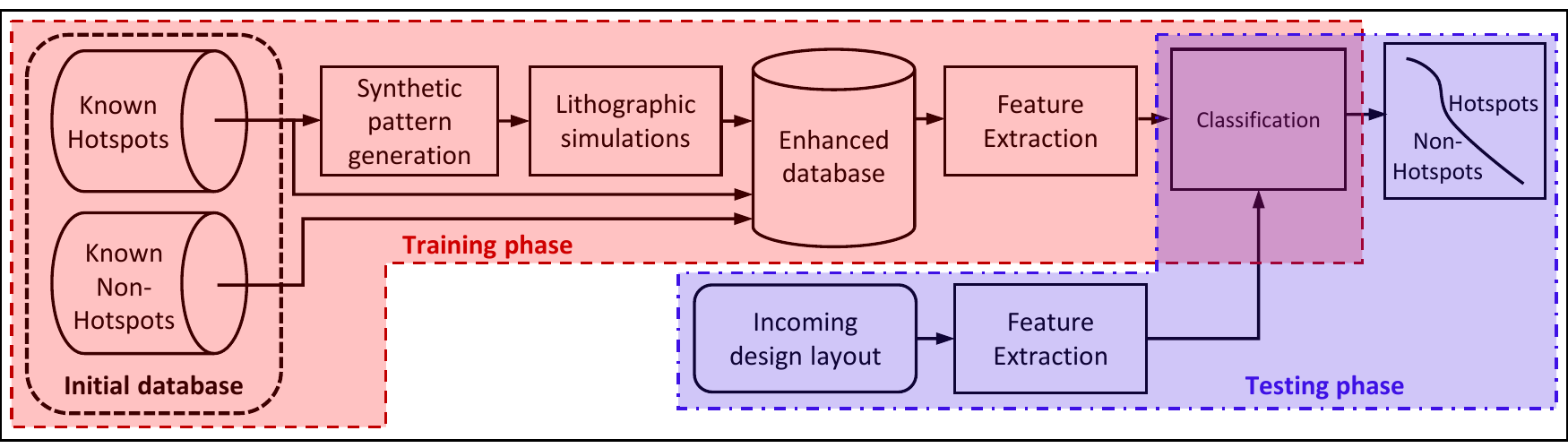}
	\vspace{-0.25in}
	\caption{The proposed machine learning-based Hotspot detection flow}
	\vspace{-0.15in}
	\label{fig:flow}
\end{figure*}

\section{Proposed Methodology} \label{sec:flow}
%I flipped the order of the two sentences below, you had them in the opposite order which did not make much sense when I read it. 

The proposed hotspot detection flow is shown in Figure \ref{fig:flow}. A high-level description is provided below and its major blocks are explained in detail in the next sub-sections. This flow is typically implemented at the foundry side and executed prior to mask fabrication; yet parts of it can be potentially incorporated into the Product Design Kits (PDKs) and transferred to the customer, in order to reduce design debug cycles.

A set of known hotspots and non-hotspots gathered from prior experience form the initial database. Design of Experiments (DOEs) is, then, performed to \textit{increase the information-theoretic content} of the initial database. As a part of these experiments, synthetic variants (patterns) of known hotspots are generated and subjected to process simulations (litho/litho-etch) to determine which of the patterns are hotspots. Synthetic patterns, along with the initial database, form the enhanced database. Patterns in the enhanced database are converted into numerical feature vectors. Feature vectors are, then, subjected to dimensionality reduction and a machine learning-based classifier (i.e., an SVM) is trained using the dimensionality-reduced feature vectors. The trained model is, then, stored to evaluate future incoming designs.
When a foundry receives a new design from its customers, it transforms it into patterns and feature vectors, and predictions are made on them using the trained classifier. Patterns classified as hotspots are subjected to further investigation, flagged as areas of interest for inline inspections, and used to drive design fixes if warranted.

\subsection{Synthetic Pattern Generation and DOEs} \label{subsec:patgen}

\begin{figure}[t]
	\centering
	%\vspace{-0.05in}
	\includegraphics[clip=true, trim=0.0in 0.4in 0.0in 0.25in, width=0.98\linewidth]{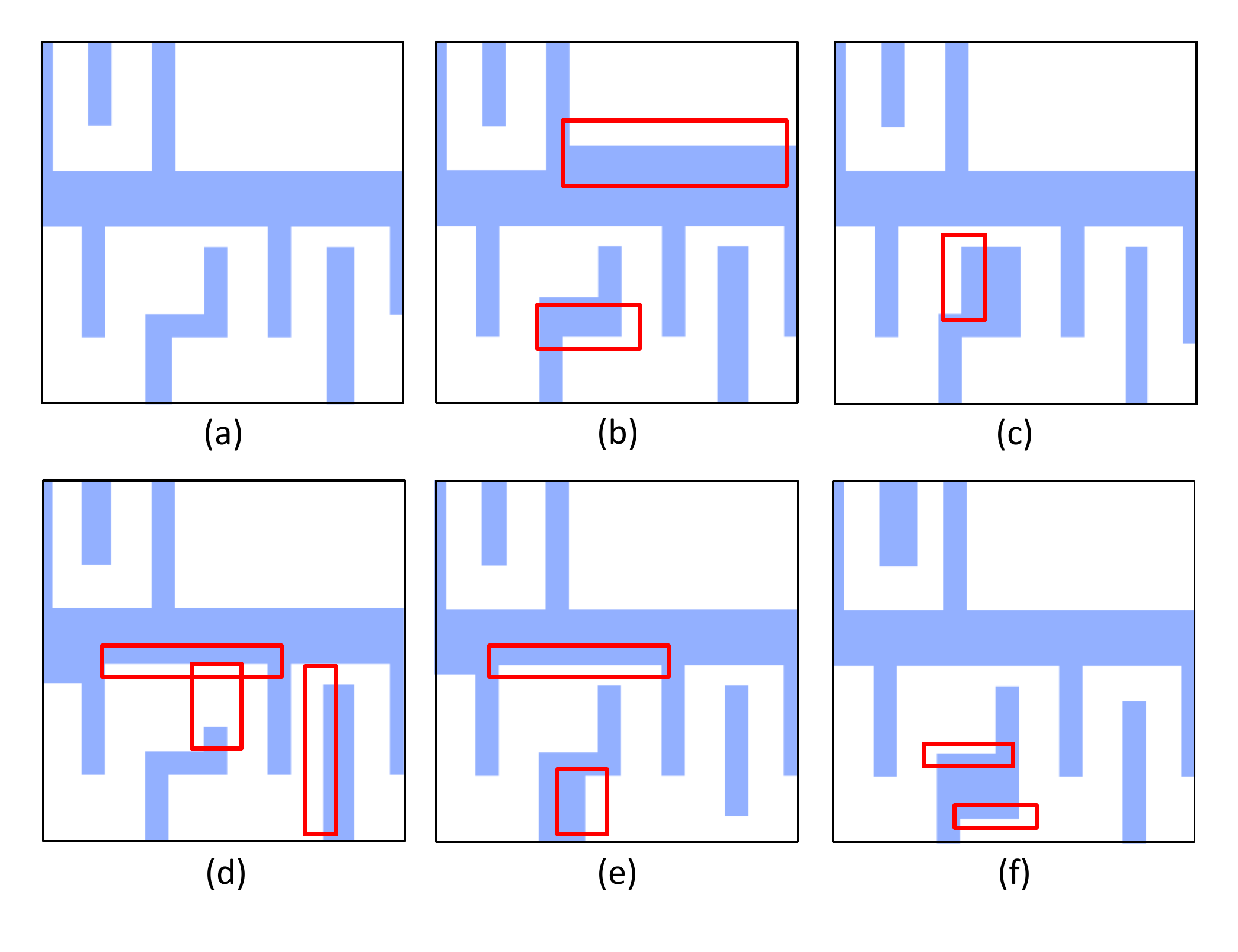}%
	\caption{(a) A hotspot pattern, (b-f) Synthetic patterns generated from pattern (a). Red markers indicate the subtle differences from pattern (a)}
	\label{fig:syn_pats}
	\vspace{-0.15in}
\end{figure}

\begin{figure*}[t!]
	\centering
	\vspace{-0.1in}
	\includegraphics[width=\textwidth, scale=0.5]{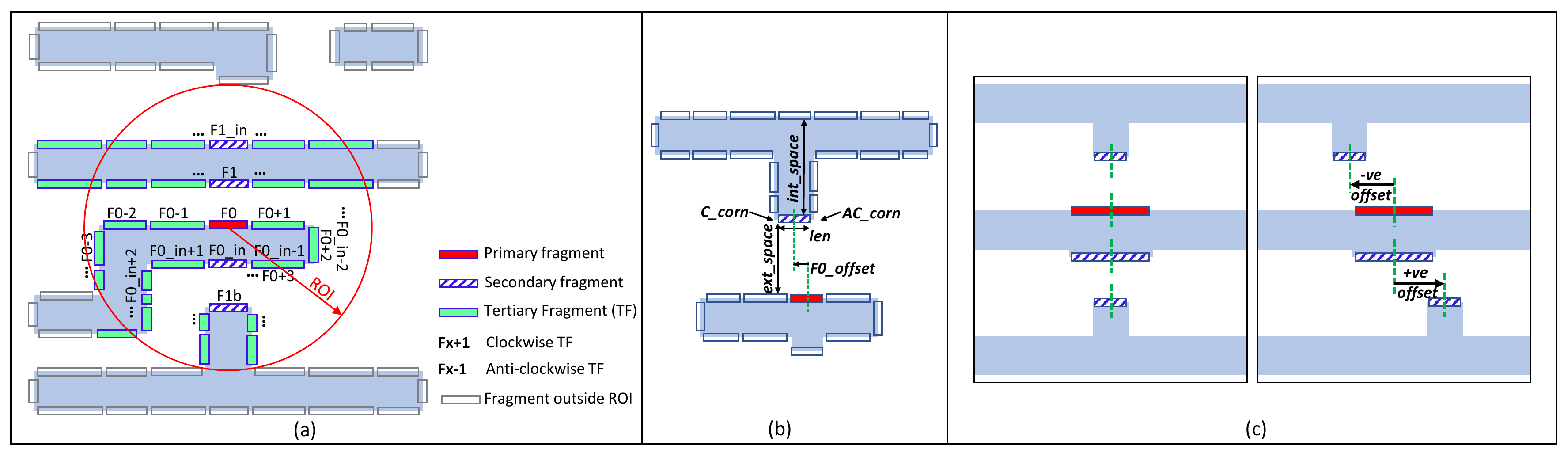}
	\vspace{-0.25in}
	\caption{Feature extraction (a) Key components of Fragment Transform Plus (FTP), (b) Fragment parameter measurements corresponding to a secondary fragment (c) Patterns illustrating the difference between FT and FTP}
	\label{fig:fxtract}
	
\end{figure*}

\begin{algorithm}[]
{\color{black}
\AlgoDontDisplayBlockMarkers\SetAlgoNoEnd%\SetAlgoNoLine%
%\footnotesize
\fontsize{8.7}{9.8}\selectfont
\LinesNumbered
\DontPrintSemicolon
\SetStartEndCondition{ }{}{}%
%\SetKwProg{Fn}{Procedure}{}{end}
\SetKwFunction{PGenerateSyntheticPatterns}{GenerateSyntheticPatterns}%
\SetKwProg{Fn}{def}{\string:}{}%
\SetKw{KwTo}{in}\SetKwFunction{Range}{range}%
\SetKwFor{For}{for}{\string:}{}%
\SetKwIF{If}{ElseIf}{Else}{if}{:}{elif}{else:}{}%
\SetKwFor{While}{while}{:}{fintq}%
\newcommand{\forcond}{polygon \KwTo HotspotPolys}
\Fn(){\PGenerateSyntheticPatterns{KnownHotspot}}{
%\KwData{Some input data\\these inputs can be displayed on several lines and one
%input can be wider than line’s width.}
\KwIn{A Known Hotspot, Synthetic pattern count, distance PDF,  Edge PDF}
\KwResult{Synthetic variants of the Hotspot}
%\tcc{this is a comment to tell you that we will now really start code}
%\If(\tcc*[h]{a simple if but with a comment on the same line}){features has power rail}{
\For{$i$ \KwTo\Range{$SynPatCount$}}{
    HotspotPolys = All polygons in the original hotspot pattern\;
    %\tcc{Add variation into POIs}
    \For{\forcond}{
        \tcc{Sample the no. of edges to be varied}
        EdgeCount = Sample from Edge PDF\;
        \For{$j$ \KwTo\Range{$EdgeCount$}}{
            \While{$EdgeAttempts \leq MaxEdges$}{
                \tcc{Randomly select an edge}
                edge = GetRandomEdge(polygon)\;
                \While{$DistAttempts<FixedCount$}{
                    dist = Sample from distance PDF\;
                    polygon = polygon.MoveEdge(edge, dist)\;
                    \tcc{Perform checks to avoid simple DRC errors}
                    MinimalDRC(ModifiedPattern)\;
                    \If(){$MinimalDRC==Pass$}{
                        go to line 5\;
                    }
                    \Else(){
                        polygon = UnmodifiedPolygon\;
                        $DistAttempts+=1$\;
                        try a different $dist$ value (go to line 8)\;
                    }
                }
                $EdgeAttempts+=1$\;
                try a different $edge$ (go to line 6)\;
            }
        }
    }
    \tcc{All polygons with/without updates, together form the modified pattern}
    SyntheticPattern = All Polygons (including modifications)\;
}
\tcc{Return patterns with variations}
\KwRet{SyntheticPatterns}
}
\vspace{0.1in}
  \caption{Synthetic pattern generation}\label{algo:patgen}
}  
\end{algorithm}

For every hotspot in the initial database, multiple synthetic patterns are generated by changing one or more features at a time. Features such as corner-to-corner distances, jogs, line-end positions, layer spacing, layer area etc., are varied. Figure \ref{fig:syn_pats} (a) shows one such hotspot and Figures \ref{fig:syn_pats} (b-f) show some of its synthetic variants. A time-efficient method for varying these features relies on perpendicularly moving the edges of one or more polygons in each snippet by a randomly sampled distance. This approach allows to quickly generate multiple patterns whose variance can be easily controlled by two parameters. The first parameter, $p$, is the probability of any given edge to move or remain stationary. By increasing this probability, we effectively increase the number of polygons and their edges that are altered in the snippets. The second parameter, $d$, is associated with a distribution of distances (Probability Density Function (PDF)), which is sampled for every polygon edge selected by the first parameter. The sampled value denotes the distance by which the edge will be displaced. These distance values follow a normal distribution centered at 0. In this way, most synthetic patterns are slight variants of the original pattern, thereby enabling us to learn the root causes effectively. However, the variation between generated patterns can be easily changed by varying the parameter $d$. Essentially, the parameter $d$ can be thought of as the standard deviation of this distribution. Parameters $p$ and $d$ are varied based on domain knowledge and experimentation. {\color{black}The pattern generation procedure is detailed in Algorithm \ref{algo:patgen}.}

As expected, the above-mentioned procedure results in a plethora of patterns, many of which might not even pass the DRC. To ensure that valid layout topologies are generated and to make this process run-time efficient, we implemented a minimal DRC engine in Python, which we execute after every pattern is generated. This check ensures that most of the generated patterns are valid. However, since implementing complex design rule checks becomes complicated, all synthetic patterns which pass this minimal DRC check are also subjected to a full DRC using CalibreDRC. Through this approach, we can ensure that the vast majority
% no need to give a number here, about 98\% 
of the generated patterns are DRC clean and usable. Synthetic patterns are, then, subjected to lithographic simulations to ascertain the ground truth about them. To this end, it is assumed that litho models are well-calibrated to the process, as is often the case in mature processes (with PDKs 1.0 and above). On the other hand, during early technology development, foundries may not have well-calibrated models readily available, but do have access to plenty of test silicon. In those situations, simulation results from crude models can be used as a guide to direct actual silicon-based experiments \cite{ReddyISTFA18}.

The number of synthetic patterns necessary to significantly improve the information-theoretic content in the training set depends on the process node, design complexity, layer of interest, etc. We have studied this dependency on a 45nm process and a detailed explanation can be found in the experimental results section. In general, these experiments are not run-time intensive, as they work with small layout snippets.  Moreover, this is a one-time procedure, hence a large number of synthetic patterns could be generated. Synthetic patterns, along with their litho simulation results, are added into the initial database in order to create the enhanced database/dataset.

\subsection{Feature Extraction} \label{subsec:fxtract}
In most of the previously proposed ML-based hotspot detection schemes, hotspot and non-hotspot patterns are initially obtained in the form of layout snippets and then subjected to Feature Extraction (FE), whereby the image snippet is transformed into a numerical feature vector which can be used to train/test a machine learning entity. In the past, various feature extraction methods, such as bounded rectangle region-based \cite{DPAN_ICICDT}, polygon fragment-based \cite{DPAN_TCAD11}, concentric circle sampling-based \cite{BEIYU_ICCAD_16}, density transform \cite{fuzzy_PM}, etc., have been proposed, suited to the detection flow they were used in. Every method has its own drawbacks and there is no clear winner among them. Density transformation is easy to implement, most widely used, and works reasonably well, but it fails to capture the minor variations between patterns which are crucial for effective learning. Co-ordinate transform \cite{ReddyVTS18} was introduced to overcome the drawbacks of density transform. It is a simple feature extraction method which accurately captures layout information and results in fewer features. Fragment transform, proposed in \cite{DPAN_TCAD11}, is a sophisticated feature extraction method which results in a small number of features but which is also fairly complex to implement.

In this work, we implement a slightly varied version of the Fragment Transform (FT) method, which we call Fragment Transform Plus (FTP). In this method, an entire layout is subjected to fragmentation using OPC tools \cite{OPCpro} and transformed into a large set of fragments as shown in Figure \ref{fig:fxtract}(a). Such an abstraction makes this feature extraction method rotate-, mirror- and flip-invariant. Post-fragmentation, every fragment is uniquely identified and analyzed individually to make hotspot/non-hotspot decisions. 

Though the decision is made at the fragment level, during the training and evaluation procedures, the context (spatial arrangement of polygons in the neighborhood) is taken into consideration. To obtain the context of a fragment, a metric called Radius Of Influence (ROI) is used. The ROI is determined by the lithography tools and the wavelength of light used for patterning. As shown in Figure \ref{fig:fxtract}(a), if a circle with radius $ROI$ is drawn by centering on the fragment under consideration, it is assumed that this circle encloses all the fragments that play a role in causing a hotspot at its center. The fragment under consideration is called the \textit{primary fragment}. {\color{black}Its parallel neighbors on either side, along a line perpendicular to its surface, are called \textit{secondary fragments}}. The lateral neighbors of both primary and secondary fragments are called \textit{tertiary fragments}. The primary fragment, along with its secondary and tertiary fragments, together can be regarded as a `pattern', similar to the patterns captured using traditional moving window-based methods \cite{ReddyVTS18}.

\textit{Fragment parameters:} To accurately capture the characteristics of a pattern, the following set of parameters are measured for every fragment within a pattern\footnote{$F0\_offset$ is not measured for primary and tertiary fragments. $ext\_space$ is not measured for some secondary and tertiary fragments which are along the periphery of a pattern. Such features are omitted because they are either redundant or they are unnecessary to accurately capture the information within a pattern.}:
\begin{equation}
\begin{split}
    fragment\_parameters =[len,\ ext\_space,\\ int\_space,\ C\_corn,\ AC\_corn,\ F0\_offset]
\end{split}
\end{equation}
where:
\begin{itemize}
    \item $len$: length of a fragment
    \item $ext\_space$: distance to the externally opposite fragment
    \item $int\_space$: distance to the internally opposite fragment
    \item $C\_corn$: corner information (convex/concave/no\_corner) from the clockwise end of the fragment
    \item $AC\_corn$: corner information from the anti-clockwise end of the fragment
    \item $F0\_offset$: offset of secondary fragments w.r.t. the location of the primary fragment
\end{itemize}
The fragment parameters for a secondary fragment of a sample pattern are depicted in Figure \ref{fig:fxtract}(b).

\textit{Feature vector generation:} The fragmentation procedure creates a different number of fragments for different patterns within the same ROI, which may result in a different number of features for every pattern. Most ML algorithms, however, expect the entire dataset to have the same number of features/dimensions. Therefore, to ensure that all patterns result in the same number of features, the ROI is abstracted as a \textit{neighboring fragment depth}. Essentially, a fixed number of perpendicularly opposite fragments, as well as lateral fragments, are considered as the neighboring fragments of the primary fragment. The depth value is chosen such that all fragments within the $ROI$ are included in the pattern. The fragment parameters of all the fragments within a pattern are, then, concatenated together to make a feature vector. For this purpose, we follow the same procedure as detailed in \cite{DPAN_TCAD11}.

\textit{Differences between FT and FTP:} To minimize the information loss during the FE procedure, as well as the total number of resultant features, we made the following changes to the original FT method:
\begin{enumerate}
    \item \textbf{A new fragment parameter $F0\_offset$ is added:} Although not apparent in \cite{DPAN_TCAD11}, the FT method fails to accurately capture the spatial arrangement of fragments which are located slightly farther from the primary fragment. For instance, the previously proposed FT method produces the same feature vector for both patterns shown in Figure \ref{fig:fxtract}(c), even though they are slightly different from each other. As noted by the authors of \cite{yang2017imbalance}, even minor $nm$-level variation could mean the difference between a pattern becoming a hotspot or a non-hotspot. Therefore, it is necessary to accurately capture such differences between patterns while performing FE. In order to avoid such \textit{transformation loss}, we introduce a new fragment parameter called $F0\_offset$. $F0\_offset$ is the offset of the center of a secondary fragment w.r.t. the center of the primary fragment, along the axis of orientation of the primary fragment. $F0\_offset$ is measured for all secondary fragments and it can be either positive or negative. Offset of the secondary fragment in the anticlockwise direction of the primary fragment is considered as negative, whereas offset in the clockwise direction is considered as positive.
    
    \item \textbf{Fragment orientation is omitted:} Fragment-based FE methods are preferred to be mirror-, flip-, and rotation-invariant. {\color{black} Such orientation-invariance assists in keeping training dataset sizes small.} The previously proposed FT method, however, includes \textit{fragment orientation} as one of its features. Including such a feature makes the FT method orientation-specific. Therefore, in order to make FTP truly rotation invariant, we omit the \textit{fragment orientation} parameter while generating feature vectors\footnote{\color{black}Orientation-invariant FE must be used only on metal layers which are patterned using symmetric illumination shapes in scanner optics. In cases of asymmetric illumination, the orientation feature must be included as part of the feature vector.}.
    
    \item \textbf{Both clockwise and anticlockwise corners of a fragment are considered:} As per \cite{DPAN_TCAD11}, it is unclear whether the FT method records information related to both corners of a fragment. In FTP, however, both clockwise and anticlockwise corner information is recorded as fragment parameters.
\end{enumerate}

\textit{Weighted features:} As an option, weights can be assigned to various fragments within a pattern. Typically, minor variations in the central area of a pattern have high influence in causing hotspots, while this influence fades as we move towards the periphery. Given enough data, an effective machine learning entity can learn this variation by itself; however, adding domain knowledge such as the aforementioned weights, helps significantly in reducing training times, as well as in increasing accuracy when working with smaller datasets.

%As an option, weights can be assigned to various fragments within a pattern, as shown in Figure \ref{fig:fxtract}d. 

\subsection{Dimensionality Reduction} \label{subsec:pca}
Dimensionality reduction algorithms reduce the number of features while retaining most of the variation in the dataset. We use Principal Component Analysis (PCA) for this purpose. PCA finds the possibly correlated features and converts them into linearly uncorrelated features called `principal components'. The benefits of working with principal components is manifold: (i) They assist in data visualization by allowing us to plot hyper-dimensional data in lower dimensions and get a better perspective of the data distribution, (ii) They help reduce ML model complexity, and (iii) They ensure smaller training times. Further information on PCA can be found in \cite{PCA}.

\subsection{Classification} \label{subsec:classify}
Hotspot detection requires a robust two-class classifier which can learn a separation boundary between hotspots and non-hotspots with maximum margin. In this work, a non-linear SVM with a Radial Basis Function (RBF) kernel is used. Detailed discussion of SVMs is out of the scope of this work; additional information can be found in \cite{svm}.

\textit{Handling imbalanced datasets:} Typically, the number of known hotspot patterns available for training is small in comparison to known non-hotspot patterns. Training with such imbalanced datasets results in a skewed classifier, which tends to favor the dominating class. Traditionally, to avoid this problem, the minority class is re-sampled (replicated) and the class sizes are equalized \cite{class_balance}. Such re-sampling, however, does not increase the information theoretic content of the dataset but only increases its size, thereby increasing model training times. In this work, we tackle this problem by setting the regularization/penalty parameter $C$ of SVMs separately for hotspots and non-hotspots. $C$ is set such that mispredictions on the minority class are penalized more in comparison to the mispredictions on the majority class. More specifically,
\begin{equation}
    C_i = class\_weight_i \cdot C
\end{equation}
\begin{equation}\label{eq:class_weight}
    class\_weight_i = \frac{total\_training\_samples}{number\_of\_classes \cdot samples_i}
\end{equation}
where, $i\ \in \ \{hotspots,\ non\_hotspots\}$

\section{Experimental Results}\label{sec:expresults}

The objective of this work is to show that enhancing the training dataset using synthetic patterns indeed increases its information-theoretic content and, in turn, reduces false-alarms. To demonstrate this, we implemented the ML-based hotspot detection flow shown in Figure \ref{fig:flow}. The classifier in this flow is trained with and without an enhanced dataset and tested against a common testing dataset. In the rest of the paper, we refer to the classifier trained with an enhanced dataset as `enhanced classifier' and to the one trained with the non-enhanced dataset as `non-enhanced classifier'. The non-enhanced classifier is the State-Of-The-Art (SOTA). The difference between the prediction results of the two classifiers indicates the effectiveness of this approach.

\subsection{Experimental Setup}\label{subsec:ExpSetup}
To generate baseline designs for our analysis, we obtained the Register-Transfer-Level (RTL) code of several designs from \cite{opencores}, synthesized, placed and routed them using the Nangate Open Cell Library (OCL) \cite{nangate}, which is based on a 45nm PDK \cite{freepdk45}. All layouts were subjected to full DRC and were found to be DRC clean. The Metal1 (M1) layer of these layouts was subjected to computation-intensive full-chip lithographic simulations using the Calibre Litho-Friendly-Design (LFD) tool-kit \cite{lfd} and the litho models provided in the PDK. These simulations ascertained the ground truth by identifying all the hotspots in the layouts. All layouts were, then, converted to patterns and feature vectors using the FTP method described in Section \ref{subsec:fxtract}. While implementing FTP, we used Calibre OPCpro \cite{OPCpro} for fragmentation and Calibre Standard Verification Rule Format (SVRF) technology for feature extraction. FTP with an ROI of $500nm$ was considered. The ROI value was abstracted as a neighboring fragment depth value of 4. After filtering out the redundant features, which are inherently created by the FTP method, the resulting dataset consisted of 519 features.

{\color{black}All of our experiments were performed on a Linux server containing an Intel Xeon E5-2660 (2.6 GHz) CPU\footnote{Experiments were performed on a server with shared resources. Therefore, run-times could slightly vary depending on other execution loads.}. All experiments used single-threaded execution except some parts of the FE procedure which was executed in parallel using up to 10 threads. The FE procedure on our smallest layout (SPI) took about 12 minutes and scaled linearly with the increase in layout area. We also note that, if necessary, larger layouts can be partitioned into smaller blocks and run in parallel, essentially yielding the same results at lower run-times.
}

%Different layouts---from the same PDK---were used for training and testing as detailed below.

\textit{Non-Enhanced Training Dataset:} In reality, to train hotspot detection models, foundries gather patterns from the first few designs manufactured in a given technology node. This dataset is usually significantly limited in size as compared to the large number of patterns which will be tested using the trained models during the lifetime of the node. To replicate such a scenario, we randomly sample two layouts and obtain a small dataset containing a total of only 100,000 patterns, which we use as our non-enhanced training dataset. Further details about this dataset, including the number of hotspots (HT) and non-hotspots (NHT), are shown in Table \ref{tab:NonEnhTrainLayouts}.

\textit{Enhanced Training Dataset:} The non-enhanced dataset generated in the previous step contains 1932 hotspots. For every one of these hotspots, we used our method to generate 500 synthetic variants. Of them, an average of approximately 484 passed DRC and, among them, about 200 were used for training. Litho simulations were performed on all DRC-clean synthetic patterns in order to obtain the ground truth (i.e., whether they are hotspots or non-hotspots). The synthetic patterns along with the patterns in the non-enhanced dataset, together form the `Enhanced Training Dataset'. Details corresponding to the enhanced training dataset are shown in Table \ref{tab:EnhTrainLayouts}.

\textit{Testing Dataset:} To mimic the real-life scenario  wherein new layouts are tested against pre-trained hotspot detection models, we evaluate the effectiveness of our method using three complete layouts which were never used during training. All patterns from these three layouts, together, comprise our testing dataset. Further details about the testing dataset are shown in Table \ref{tab:TestLayouts}.

\begin{table}[t!]
\caption{Non-Enhanced Training dataset}
\begin{tabular}{|l|c|c|c|c|}
\hline
\multicolumn{1}{|c|}{\textbf{Layout}} & \textbf{Size} & \textbf{Sample Count} & \textbf{HT\#} & \textbf{NHT\#} \\ \hline
wb\_conmax                            & \SI{205}{\micro\meter} $\times$ \SI{205}{\micro\meter}   & 50,000                & 1,236          & 48,764 \\ \hline
Ethernet                              & \SI{360}{\micro\meter} $\times$ \SI{360}{\micro\meter}   & 50,000                & 696           & 49,304 \\ \hline
Total                              & Not applicable   & 100,000                & 1,932           & 98,068 \\ \hline
\end{tabular}
\label{tab:NonEnhTrainLayouts}
\end{table}
%\SI{171625}{\micro\meter\squared} -- total area

\begin{table}[t!]
\centering
\caption{Enhanced Training dataset}
\begin{tabular}{|l|c|c|c|}
\hline
\textbf{Pattern Type} & \textbf{Pattern Count} & \textbf{HT\#} & \textbf{NHT\#} \\ \hline
Non-Enhanced Dataset  & 100,000                 & 1,932         & 98,068         \\ \hline
Synthetic patterns    & 386,136                 & 192,680       & 193,456        \\ \hline
Total    & 486,136                 & 194,612       & 291,524        \\ \hline
\end{tabular}
\label{tab:EnhTrainLayouts}
\end{table}

\begin{table}[t!]
\caption{Testing dataset}
\begin{tabular}{|l|c|c|c|c|}
\hline
\multicolumn{1}{|c|}{\textbf{Layout}} & \textbf{Size}  & \textbf{\begin{tabular}[c]{@{}c@{}}Pattern\\ Count\end{tabular}} & \textbf{HT\#} & \textbf{NHT\#} \\ \hline
SPI                                   & \SI{66}{\micro\meter} x \SI{66}{\micro\meter}      & 326,394                & 5,768         & 320,626        \\ \hline
TV80                                  & \SI{96}{\micro\meter} x \SI{96}{\micro\meter}      & 789,253                & 16,736        & 772,517        \\ \hline
AES (Encrypt)                                  & \SI{160}{\micro\meter} x \SI{160}{\micro\meter}    & 1,999,419              & 47,700        & 1,951,719      \\ \hline
HTC patterns                          & Not applicable & 384,736            & 191,345        & 193,391         \\ \hline
Total                          & Not applicable & 3,499,802            & 261,549        & 3,238,253         \\ \hline
\end{tabular}
\label{tab:TestLayouts}
\end{table}

\textit{Hard-To-Classify Test Patterns:} The authors of \cite{dai2017optimization} have performed an interesting study, wherein they compare the layout patterns used during Technology Development (TD) against the patterns found in product designs. They discovered that, while some patterns in the product designs were topologically similar to the patterns seen during TD, the product designs had many more dimensional variations of those patterns. In a separate study, according to the observations of the authors in \cite{yang2017imbalance}, even minor variations in the widths and spaces between polygons of a pattern can mean the difference between a pattern becoming a hotspot or a non-hotspot.
If we extend the results of these studies to hotspot detection, we can envision a scenario wherein a hotspot detection model is trained using a certain hotspot but tested with many more variations of that same hotspot, which could be either hotspots or non-hotspots. For example, if the pattern shown in Figure \ref{fig:case_study}(a) was used in training, then many more variations of this pattern, such as the ones shown in Figures \ref{fig:case_study}(b--d) --which are non-hotspots-- could be found in future product designs and tested using the trained model. Such patterns, as described in Section \ref{subsec:falarms}, are the real source of false alarms as they lie in close proximity to the training hotspots in the hyper-dimensional space and are truly hard-to-classify.
\begin{figure}[t!]
	\centering
	\vspace{-0.125in}
	\includegraphics[clip=true, trim=0.9in 0.0in 0.9in 0.0in,width=0.90\linewidth]{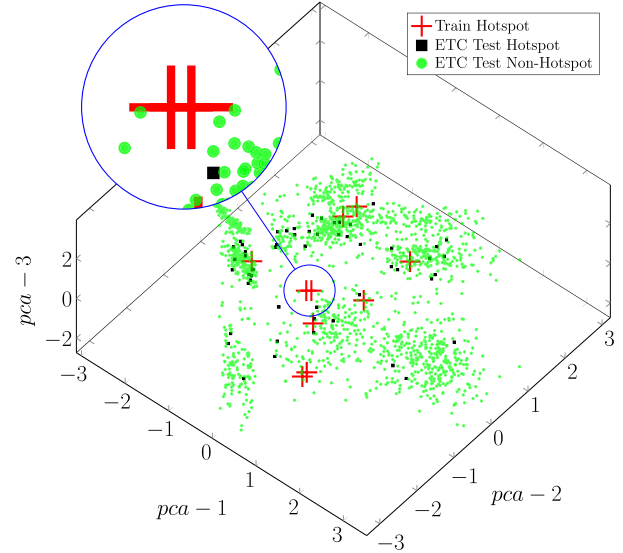}
	\vspace{-0.0in}
	\color{black}\caption{Distribution of ETC test patterns w.r.t. training hotspots}
	\label{fig:pca_ETC}
	\vspace{-0.095in}
\end{figure}

Upon detailed analysis, we found that the test layouts (i.e., the designs listed in the first three rows of Table \ref{tab:TestLayouts}), do not contain any such HTC patterns within them. Therefore, to replicate a scenario witnessed by an actual foundry, we further expanded the testing dataset by adding, for each known hotspot, approximately 200 synthetic patterns, which were never used during training. These act as HTC patterns in the testing dataset and assist in determining whether the trained model is truly robust in preventing false alarms. Details about these patterns are shown in the fourth row of Table \ref{tab:TestLayouts}. In the rest of the paper, patterns other than the HTC patterns (i.e., the first three rows of Table \ref{tab:TestLayouts}), are referred to as Easy-To-Classify (ETC) patterns. 

\begin{figure}[t]
	\centering
	\vspace{-0.125in}
	\includegraphics[clip=true, trim=0.9in 0.0in 0.9in 0.0in,width=0.90\linewidth]{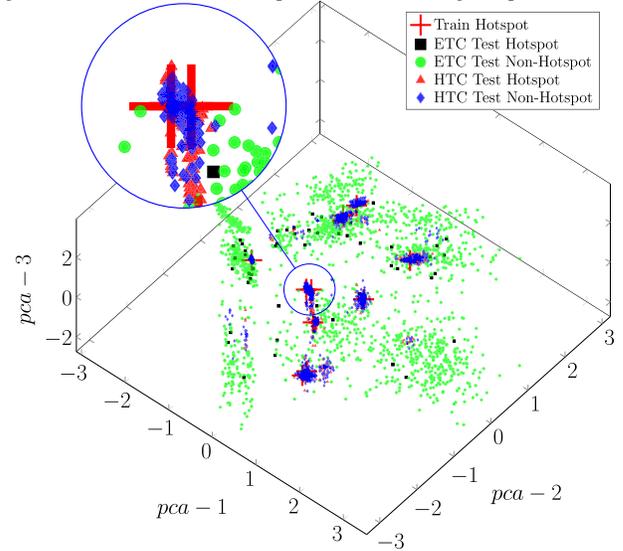}
	\vspace{-0.0in}
	\color{black}\caption{Distribution of ETC and HTC test patterns w.r.t. training hotspots}
	\label{fig:pca_HTC}
	\vspace{-0.095in}
\end{figure}

To further demonstrate the importance of including HTC patterns in the testing dataset and to contrast their distribution against the ETC patterns, we perform PCA on the training dataset and project both the ETC and the HTC patterns onto the same space. For the sake of brevity, we plot only 10 randomly sampled hotspots from the training dataset, their corresponding HTC test patterns (2000 in total), and the same number of randomly sampled ETC test patterns. Figure \ref{fig:pca_ETC} shows the distribution of just the ETC test patterns w.r.t. to the training hotspots. Figure \ref{fig:pca_HTC} is similar to Figure \ref{fig:pca_ETC} but it also includes the HTC test patterns. By contrasting the two figures, we observe that HTC test non-hotspots are located in much closer proximity to the training hotspots, when compared to the proximity of ETC test non-hotspots to the training hotspots. Therefore, HTC test non-hotspots are prone to be misclassified as hotspots. Such nature of HTC patterns makes their presence in the testing dataset essential for accurately evaluating the effectiveness of hotspot detection methods. 

\subsection{Classifier and hyper-parameter selection}
As explained in Section \ref{subsec:classify}, an SVM with an RBF kernel is used as a two-class classifier. While training a hotspot detection model, we can trade off its accuracy and its false-alarm rate, thereby setting the operating region of the trained model. For instance, a model can be tuned to achieve high accuracy rates by allowing a slight increase in false alarms, or it can be tuned to achieve very low false alarm rates with a small reduction in accuracy. Such adjustments in the performance of a classifier, or any other ML entity, can be made by tuning its hyper-parameters. The commonly tuned hyper-parameters of an SVM with an RBF kernel are $C$ and $gamma$. While the $C$ parameter trades off the classification accuracy of training examples against maximization of the decision function margin, the $gamma$ parameter controls the extent of influence of a single training example. Also, in order to bias the classifier towards a particular class, we can add bias factors to the $class\_weight$ parameter shown in Equation \ref{eq:class_weight}.

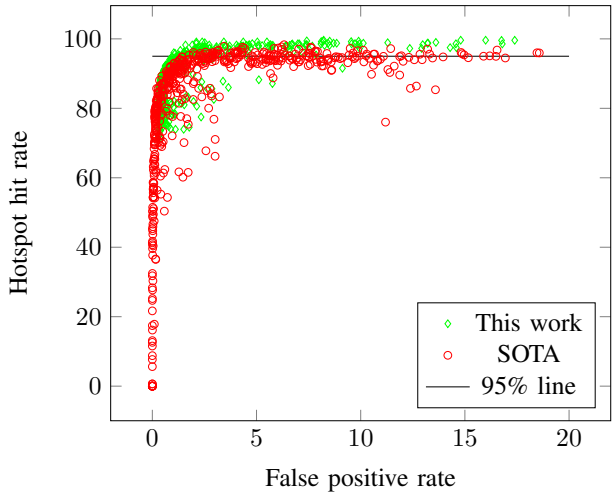
\begin{figure}[t!]
	\vspace{-0.05in}
	\centering
	\begin{tikzpicture}[scale=0.97]
    \begin{axis}[
        legend pos=south east,
        xlabel=False positive rate,
        ylabel=Hotspot hit rate,
    ]
    
    \addplot[
        only marks,
        mark=diamond,
        mark size=1.5pt,
        color=green]
    table [x=ours_fprate, y=ours_hitrate, col sep=comma] 
    {Figures/ETC_plot_ours_data.csv};
    \addplot[
        only marks,
        mark=o,
        mark size=1.5pt,
        color=red]
    table [x=sota_fprate, y=sota_hitrate, col sep=comma] 
    {Figures/ETC_plot_sota_data.csv};
    \addplot [
    domain=0:20, 
    samples=100, 
    color=black,
]
{95};
    \legend{This work,SOTA,95\% line}
    \end{axis}
    \end{tikzpicture}
    \caption{Hyper-parameter tuning analysis using ETC test patterns}
    \label{fig:hyper_param_sweep_ETC}
\end{figure}

%Bias factors can be added to the $class\_weight$ parameter shown in Equation \ref{eq:class_weight} in order to bias the classifier towards a particular class.

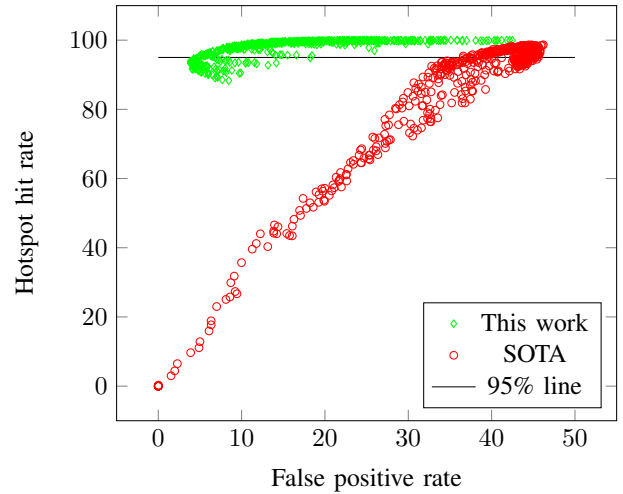
\begin{figure}[]
	\vspace{-0.05in}
	\centering
	\begin{tikzpicture}[scale=0.97]
    \begin{axis}[legend pos=south east,
        xlabel=False positive rate,
        ylabel=Hotspot hit rate,]
    \addplot[
        only marks,
        mark=diamond,
        mark size=1.5pt,
        color=green]
    table [x=ours_fprate, y=ours_hitrate, col sep=comma] 
    {Figures/HTC_plot_ours_data.csv};
    \addplot[
        only marks,
        mark=o,
        mark size=1.5pt,
        color=red]
    table [x=sota_fprate, y=sota_hitrate, col sep=comma] 
    {Figures/HTC_plot_sota_data.csv};
        \addplot [
    domain=0:50, 
    samples=100, 
    color=black,
]
{95};
    \legend{This work,SOTA, 95\% line}
    \end{axis}
    \end{tikzpicture}
        
    \caption{Hyper-parameter tuning analysis using HTC test patterns}
\label{fig:hyper_param_sweep_HTC}
\end{figure}

In this work, we used the grid search method \cite{syarif2016gridsearch}\cite{GCV} and 3-fold cross validation \cite{Kohavi1995modelselect} for both non-enhanced classifiers (SOTA) and enhanced classifiers (proposed) to choose the optimal hyper-parameters. We randomly sampled 75,000 patterns from the training dataset and used this dataset for tuning and cross-validation. We, then, augmented this dataset with about 200 synthetic patterns per hotspot while tuning the enhanced classifier. For both classifiers, we swept the parameters $C$, $gamma$, and $class\_weight$ across a wide range. The results of this analysis are shown in Figures \ref{fig:hyper_param_sweep_ETC} and \ref{fig:hyper_param_sweep_HTC}. Figure \ref{fig:hyper_param_sweep_ETC} shows a scatter plot which demonstrates the effect of hyper-parameters on the performance of the classifier when tested using ETC patterns. This plot contains 546 data-points for each of the enhanced and non-enhanced classifiers. Each data-point shows the hotspot hit rate (accuracy) and false positive (false alarm) rate observed for a certain combination of hyper-parameter values. While such a grid search analysis shows several sub-optimal combinations, the pareto front of the scatter plot indicates the optimal hyper-parameter combinations. Depending on the desired region of operation, the user can choose any point along the pareto front, essentially trading off accuracy and false-alarm rate. Figure \ref{fig:hyper_param_sweep_HTC} shows the results from a similar analysis performed by using only HTC patterns as test patterns.

Comparing Figures \ref{fig:hyper_param_sweep_ETC} and \ref{fig:hyper_param_sweep_HTC}, we observe that both classifiers (i.e., non-enhanced and enhanced) perform similarly on ETC test patterns; however, when tested using HTC patterns, we observe that the SOTA (i.e., non-enhanced classifier) suffers from very high false alarm rates and provides sub-optimal results across all hyper-parameter combinations, an issue that is clearly resolved by the proposed method (i.e., enhanced classifier).

% Table generated by Excel2LaTeX from sheet 'Sheet1'
\begin{table}[b]
	\centering
	\vspace{-0.1in}
	\caption{Metrics and Formulas}
    \vspace{-0.2in}
	\begin{threeparttable}
	\hspace*{-0.5cm}\begin{tabular}{l c}%{ m{3cm} m{4cm}  }
		\toprule
		\multicolumn{1}{c}{\textbf{Metric}} & \multicolumn{1}{c}{\textbf{Formula Used}} \\
		\midrule
		Hotspot (HT) hit rate & $\frac{predicted\_hotspots}{total\_hotspots}$ \\
		%\midrule
		Non-Hotspot (NHT) hit rate  & $\frac{predicted\_non\_hotspots}{total\_non\_hotspots}$ \\
		%\toprule
		False Positive (FP) rate & $\frac{false\_pos}{total\_patterns\_tested}$ \\
		%\midrule
		False Negative (FN) rate & $\frac{false\_neg}{total\_patterns\_tested}$ \\
		%\midrule
		Total error rate & $\frac{false\_pos + false\_neg}{total\_patterns\_tested}$ \\
		\color{black}Matthews Corr. Coeff. (MCC) & \color{black}$\frac{TP \times TN - FP \times FN}{\sqrt{(TP+FP)(TP+FN)(TN+FP)(TN+FN)}}$ \\
		\bottomrule
	\end{tabular}
  \begin{tablenotes}
    \color{black}\item[] TP: True Positives (same as $predicted\_hotspots$)
    \color{black}\item[] TN: True Negatives (same as $predicted\_non\_hotspots$)
  \end{tablenotes}
\end{threeparttable}

	\label{tab:formulas}
	%\vspace{-0.2in}
\end{table}

\subsection{Experimental Analysis}\label{subsec:exp_analysis}
In order to delve into a more detailed analysis, we need to choose a specific operating point for both the enhanced and the non-enhanced classifiers along their pareto fronts shown in Figure \ref{fig:hyper_param_sweep_ETC}. While a user could choose any point of interest, for the purpose of this analysis we chose the operating point which is closest to the 95\% accuracy rate for each of the two classifiers.

For these points, we take the corresponding hyper-parameters for the non-enhanced and the enhanced classifiers, we train them using the training datasets shown in Tables \ref{tab:NonEnhTrainLayouts} and \ref{tab:EnhTrainLayouts}, respectively, and we test them using a common testing dataset. The formulas for the various metrics used in our analysis are shown in Table \ref{tab:formulas}. We test both the non-enhanced and the enhanced classifiers using the testing dataset and the results for ETC patterns and HTC patterns are shown in Tables \ref{tab:res_withpca_ETC} and \ref{tab:res_withpca_HTC}, respectively. From these results, we observe that both the SOTA and the proposed method perform similarly well on ETC test patterns, by showing average hotspot hit rates of about 89\% and false positive rates less than 2\%. In the case of HTC patterns, however, the SOTA shows very high false positives. The proposed method, on the other hand, reduces false positives by about 69\% (\% change from 40.42\% to 12.41\%) in comparison to the SOTA.

%By comparing the results with and without dimensionality reduction, we observe that the dimensionality reduction procedure does not introduce any significant error into our analysis, while at the same time it provides various benefits, as explained in Section \ref{subsec:pca}.

\begin{table}[b]
\color{black}
\centering
\caption{Runtime information}

\hspace*{-0.2cm}\begin{tabular}{|l|c|c|c|c|c|c|}
\hline
\multicolumn{1}{|c|}{\multirow{2}{*}{\textbf{\begin{tabular}[c]{@{}c@{}}Classifier\\   Type\end{tabular}}}} & \multirow{2}{*}{\textbf{\begin{tabular}[c]{@{}c@{}}Training\\ Time\end{tabular}}} & \multicolumn{5}{c|}{\textbf{Testing  Time}} \\ \cline{3-7} 
\multicolumn{1}{|c|}{} &  & \textbf{SPI} & \textbf{TV80} & \textbf{AES} & \textbf{\begin{tabular}[c]{@{}c@{}}HTC\\ Patterns\end{tabular}} & \textbf{\begin{tabular}[c]{@{}c@{}}Average\\ (/pat.)\end{tabular}} \\ \hline
\begin{tabular}[c]{@{}l@{}}Non-\\ enhanced\end{tabular} & 0.11h & 1.85h & 0.71h & 1.78h & 0.35h & \textbf{3.26ms} \\ \hline
Enhanced & 10.94h & \multicolumn{1}{l|}{3.02h} & \multicolumn{1}{l|}{7.11h} & \multicolumn{1}{l|}{18.48h} & 3.53h & \textbf{32.99ms} \\ \hline
\end{tabular}

\label{tab:runtime_info}
\end{table} 

{\color{black}Matthews Correlation Coefficient (MCC) is an effective indicator of the quality of two-class classification, especially when dealing with imbalanced datasets. The MCC value ranges from -1 to +1, indicating the quality of prediction from random to perfect, respectively. In our analysis, the MCC value for the enhanced classifier is about 0.7 for predictions on ETC patterns and about 0.77 for predictions on HTC patterns, confirming their high accuracy.}

\begin{table*}[]
\centering
\caption{Test results for ETC patterns (with dimensionality reduction)}

\begin{tabular}{|l|c|c|c|c|c|c|c|c|c|c|c|c|}
\hline
\multicolumn{1}{|c|}{\multirow{2}{*}{\textbf{\begin{tabular}[c]{@{}c@{}}Test\\ Layout\end{tabular}}}} & \multicolumn{6}{c|}{\textbf{SOTA (Non-enhanced)}} & \multicolumn{6}{c|}{\textbf{This work (Enhanced)}} \\ \cline{2-13} 
\multicolumn{1}{|c|}{} & \textbf{\begin{tabular}[c]{@{}c@{}}HT hit\\ rate (\%)\end{tabular}} & \textbf{\begin{tabular}[c]{@{}c@{}}NHT hit\\ rate (\%)\end{tabular}} & \textbf{\begin{tabular}[c]{@{}c@{}}FP rate\\ (\%)\end{tabular}} & \textbf{\begin{tabular}[c]{@{}c@{}}FN rate\\ (\%)\end{tabular}} & \textbf{\begin{tabular}[c]{@{}c@{}}Total Err.\\ rate (\%)\end{tabular}} & \textbf{MCC} & \textbf{\begin{tabular}[c]{@{}c@{}}HT hit\\ rate (\%)\end{tabular}} & \textbf{\begin{tabular}[c]{@{}c@{}}NHT hit\\ rate (\%)\end{tabular}} & \textbf{\begin{tabular}[c]{@{}c@{}}FP rate\\ (\%)\end{tabular}} & \textbf{\begin{tabular}[c]{@{}c@{}}FN rate\\ (\%)\end{tabular}} & \textbf{\begin{tabular}[c]{@{}c@{}}Total Err.\\ rate (\%)\end{tabular}} & \textbf{MCC} \\ \hline
SPI & 91.71 & 98.47 & 1.50 & 0.15 & 1.65 & 0.68 & 92.27 & 99.10 & 0.88 & 0.14 & 1.02 & 0.77 \\ \hline
TV80 & 86.97 & 97.92 & 2.04 & 0.28 & 2.31 & 0.63 & 88.61 & 98.01 & 1.95 & 0.24 & 2.18 & 0.65 \\ \hline
AES & 87.27 & 98.03 & 1.92 & 0.30 & 2.22 & 0.66 & 88.30 & 98.02 & 1.93 & 0.28 & 2.21 & 0.67 \\ \hline
\textbf{Average} & \textbf{88.65} & \textbf{98.14} & \textbf{1.82} & \textbf{0.24} & \textbf{2.06} & \textbf{0.66} & \textbf{89.73} & \textbf{98.38} & \textbf{1.59} & \textbf{0.22} & \textbf{1.80} & \textbf{0.70} \\ \hline
\end{tabular}

\label{tab:res_withpca_ETC}
\end{table*}

\begin{table*}[]
\centering
\caption{Test results for HTC patterns (with dimensionality reduction)}
\hspace*{-0.3cm}\begin{tabular}{|c|c|c|c|c|c|c|c|c|c|c|c|c|}
\hline
\multirow{2}{*}{\textbf{\begin{tabular}[c]{@{}c@{}}Test\\ Layout\end{tabular}}} & \multicolumn{6}{c|}{\textbf{SOTA (Non-enhanced)}} & \multicolumn{6}{c|}{\textbf{This work (Enhanced)}} \\ \cline{2-13} 
 & \textbf{\begin{tabular}[c]{@{}c@{}}HT hit\\ rate (\%)\end{tabular}} & \textbf{\begin{tabular}[c]{@{}c@{}}NHT hit\\ rate (\%)\end{tabular}} & \textbf{\begin{tabular}[c]{@{}c@{}}FP rate\\ (\%)\end{tabular}} & \textbf{\begin{tabular}[c]{@{}c@{}}FN rate\\ (\%)\end{tabular}} & \textbf{\begin{tabular}[c]{@{}c@{}}Tot. Err.\\ rate (\%)\end{tabular}} & \textbf{MCC} & \textbf{\begin{tabular}[c]{@{}c@{}}HT hit\\ rate (\%)\end{tabular}} & \textbf{\begin{tabular}[c]{@{}c@{}}NHT hit\\ rate (\%)\end{tabular}} & \textbf{\begin{tabular}[c]{@{}c@{}}FP rate\\ (\%)\end{tabular}} & \textbf{\begin{tabular}[c]{@{}c@{}}FN rate\\ (\%)\end{tabular}} & \textbf{\begin{tabular}[c]{@{}c@{}}Tot. Err.\\ rate (\%)\end{tabular}} & \textbf{MCC} \\ \hline
\multicolumn{1}{|l|}{HTC Patterns} & 97.78 & 19.58 & 40.42 & 1.10 & 41.52 & 0.28 & 99.27 & 75.31 & 12.41 & 0.36 & 12.77 & 0.77 \\ \hline
\end{tabular}
\label{tab:res_withpca_HTC}
\end{table*}

\begin{table*}[]
\centering
\caption{Test results for ETC patterns (without dimensionality reduction)}
\begin{tabular}{|l|c|c|c|c|c|c|c|c|c|c|c|c|}
\hline
\multicolumn{1}{|c|}{\multirow{2}{*}{\textbf{\begin{tabular}[c]{@{}c@{}}Test\\ Layout\end{tabular}}}} & \multicolumn{6}{c|}{\textbf{SOTA (Non-enhanced)}} & \multicolumn{6}{c|}{\textbf{This work (Enhanced)}} \\ \cline{2-13} 
\multicolumn{1}{|c|}{} & \textbf{\begin{tabular}[c]{@{}c@{}}HT hit\\ rate (\%)\end{tabular}} & \textbf{\begin{tabular}[c]{@{}c@{}}NHT hit\\ rate (\%)\end{tabular}} & \textbf{\begin{tabular}[c]{@{}c@{}}FP rate\\ (\%)\end{tabular}} & \textbf{\begin{tabular}[c]{@{}c@{}}FN rate\\ (\%)\end{tabular}} & \textbf{\begin{tabular}[c]{@{}c@{}}Total Err.\\ rate (\%)\end{tabular}} & \textbf{MCC} & \textbf{\begin{tabular}[c]{@{}c@{}}HT hit\\ rate (\%)\end{tabular}} & \textbf{\begin{tabular}[c]{@{}c@{}}NHT hit\\ rate (\%)\end{tabular}} & \textbf{\begin{tabular}[c]{@{}c@{}}FP rate\\ (\%)\end{tabular}} & \textbf{\begin{tabular}[c]{@{}c@{}}FN rate\\ (\%)\end{tabular}} & \textbf{\begin{tabular}[c]{@{}c@{}}Total Err.\\ rate (\%)\end{tabular}} & \textbf{MCC} \\ \hline
SPI & 91.47 & 98.56 & 1.41 & 0.15 & 1.56 & 0.69 & 92.29 & 99.16 & 0.82 & 0.14 & 0.96 & 0.78 \\ \hline
TV80 & 86.21 & 98.04 & 1.92 & 0.29 & 2.21 & 0.64 & 88.56 & 98.10 & 1.86 & 0.24 & 2.10 & 0.66 \\ \hline
AES & 86.50 & 98.15 & 1.81 & 0.32 & 2.13 & 0.67 & 88.04 & 98.11 & 1.85 & 0.29 & 2.13 & 0.67 \\ \hline
\textbf{Average} & \textbf{88.06} & \textbf{98.25} & \textbf{1.71} & \textbf{0.25} & \textbf{1.97} & \textbf{0.67} & \textbf{89.63} & \textbf{98.46} & \textbf{1.51} & \textbf{0.22} & \textbf{1.73} & \textbf{0.70} \\ \hline
\end{tabular}
\label{tab:res_withoutpca_ETC}
\end{table*}

\begin{table*}[t!]
\centering
\caption{Test results for HTC patterns (without dimensionality reduction)}
 \hspace*{-0.3cm}\begin{tabular}{|c|c|c|c|c|c|c|c|c|c|c|c|c|}
\hline
\multirow{2}{*}{\textbf{\begin{tabular}[c]{@{}c@{}}Test\\ Layout\end{tabular}}} & \multicolumn{6}{c|}{\textbf{SOTA (Non-enhanced)}} & \multicolumn{6}{c|}{\textbf{This work (Enhanced)}} \\ \cline{2-13} 
 & \textbf{\begin{tabular}[c]{@{}c@{}}HT hit\\ rate (\%)\end{tabular}} & \textbf{\begin{tabular}[c]{@{}c@{}}NHT hit\\ rate (\%)\end{tabular}} & \textbf{\begin{tabular}[c]{@{}c@{}}FP rate\\ (\%)\end{tabular}} & \textbf{\begin{tabular}[c]{@{}c@{}}FN rate\\ (\%)\end{tabular}} & \textbf{\begin{tabular}[c]{@{}c@{}}Tot. Err.\\ rate (\%)\end{tabular}} & \textbf{MCC} & \textbf{\begin{tabular}[c]{@{}c@{}}HT hit\\ rate (\%)\end{tabular}} & \textbf{\begin{tabular}[c]{@{}c@{}}NHT hit\\ rate (\%)\end{tabular}} & \textbf{\begin{tabular}[c]{@{}c@{}}FP rate\\ (\%)\end{tabular}} & \textbf{\begin{tabular}[c]{@{}c@{}}FN rate\\ (\%)\end{tabular}} & \textbf{\begin{tabular}[c]{@{}c@{}}Tot. Err.\\ rate (\%)\end{tabular}} & \textbf{MCC} \\ \hline
\multicolumn{1}{|l|}{HTC Patterns} & 97.64 & 20.14 & 40.14 & 1.18 & 41.32 & 0.28 & 99.27 & 75.85 & 12.14 & 0.36 & 12.50 & 0.77 \\ \hline
\end{tabular}
\label{tab:res_withoutpca_HTC}
\vspace{-0.1in}
\end{table*}

{\color{black}The training and testing run-times are shown in Table \ref{tab:runtime_info}. Evidently, both of these times appear to be slightly higher for the enhanced classifier  than for the non-enhanced classifier. However, when assessing these times, the following points should be taken into account: (i) training is a one-time procedure, hence a time-increase of the magnitude experienced herein is rather inconsequential, (ii) testing is a highly parallelizable procedure, hence the time-increase experienced herein is not a show-stopper, and (iii) the significant improvement in the quality of the trained model outweighs, by far, the slight increase required in computational time. 

%We observe an increased training time for the enhanced classifier because of the increase in the dataset size, and finer learning. Although the training time for the enhanced classifier is higher than the non-enhanced classifier, we note that this is a one-time procedure and the long-term benefits in terms of improved results outweighs the slightly increased computational costs. Similarly, testing times are also higher but this process is highly parallelizable and scales linearly.
}

The results shown in Tables \ref{tab:res_withpca_ETC} and \ref{tab:res_withpca_HTC} were obtained using dimensionality-reduced datasets. Specifically, as explained in Section \ref{subsec:pca}, PCA was used for dimensionality reduction and only the first 250 principal components were used for training and testing. To verify whether PCA introduces any additional error into the analysis, we repeated the experiments without performing dimensionality reduction (i.e., using all 519 features). The results from these experiments are shown in Tables \ref{tab:res_withoutpca_ETC} and \ref{tab:res_withoutpca_HTC}. By comparing the corresponding results with and without dimensionality reduction, we observe that the difference across all metrics is less than 0.6\%. Thereby, these results demonstrate that the dimensionality reduction procedure does not introduce any significant error into our analysis, while at the same time it provides various benefits, as explained in Section \ref{subsec:pca}.

% Preamble: \pgfplotsset{width=7cm,compat=newest}
 \begin{figure}[]
	\vspace{-0.05in}
	\centering
	\pgfplotsset{width=1.12*\linewidth, height=6cm, compat=newest}
	\pgfplotsset{every axis/.append style={
			font=\footnotesize,
			line width=1.1pt,
			tick style={line width=0.6pt}}}
	%[width=0.5*\textwidth]
	%\hspace{-0.5in}
	\hspace*{-0.2cm}\begin{tikzpicture}
	\begin{axis}[
	xlabel=Synthetic patterns per known hotspot,
	ylabel=Total Error \%,
	every axis y label/.style=
	{at={(ticklabel cs:0.5)},rotate=90,anchor=near ticklabel},
	nodes near coords,
	point meta=explicit symbolic,
	xtick=data,
	grid=major]
    \addplot+[smooth][color=blue, mark size=2pt, line width=1pt,
        mark=*,
        x=s,
        y=F,
        nodes near coords,
        point meta=explicit symbolic,
        visualization depends on={value \thisrow{anchor}\as\myanchor},,
        every node near coord/.append style={font=\footnotesize,anchor=\myanchor}
        %nodes={font=\small},
    %    nodes near coords align={anchor=west},
    ] table [row sep=\\,meta=Label]
    {
    s F Label anchor\\
		0 42.75   {42.75} west\\
		40 30.14  {30.14} west\\
		80 18.98  {18.98} 220\\
		120 15.78 {15.78} 220\\
		160 14.27 {14.27} 50\\
		200 13.17 {13.17} 270\\
		240 12.63 {12.63} 60\\
		280 12.03 {12.03} 260\\
		320 11.56 {11.56} 90\\
		360 11.23 {11.23} 270\\
		400 10.87 {10.87} 90\\
		440 10.62 {10.62} 270\\
		480 10.35 {10.35} 110\\
		520 10.17 {10.17} 240\\
    };
    \
	\end{axis}
	\end{tikzpicture}
	\vspace{-0.2in}
	\color{black}\caption{Variation of prediction error w.r.t. number of synthetic patterns used}
	\label{fig:error_var}
	\vspace{-0.2in}
\end{figure}
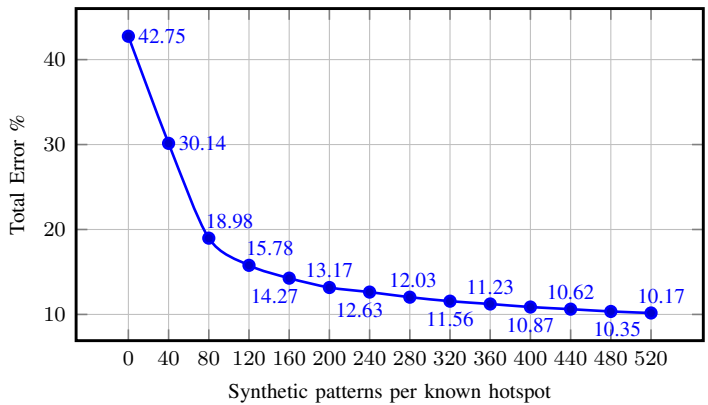

\begin{figure*}[t!]
	\centering
	\vspace{0.1in}
	\includegraphics[clip=true, trim=0.0in 0.0in 0.0in 0.0in,width=\textwidth, scale=1.0]{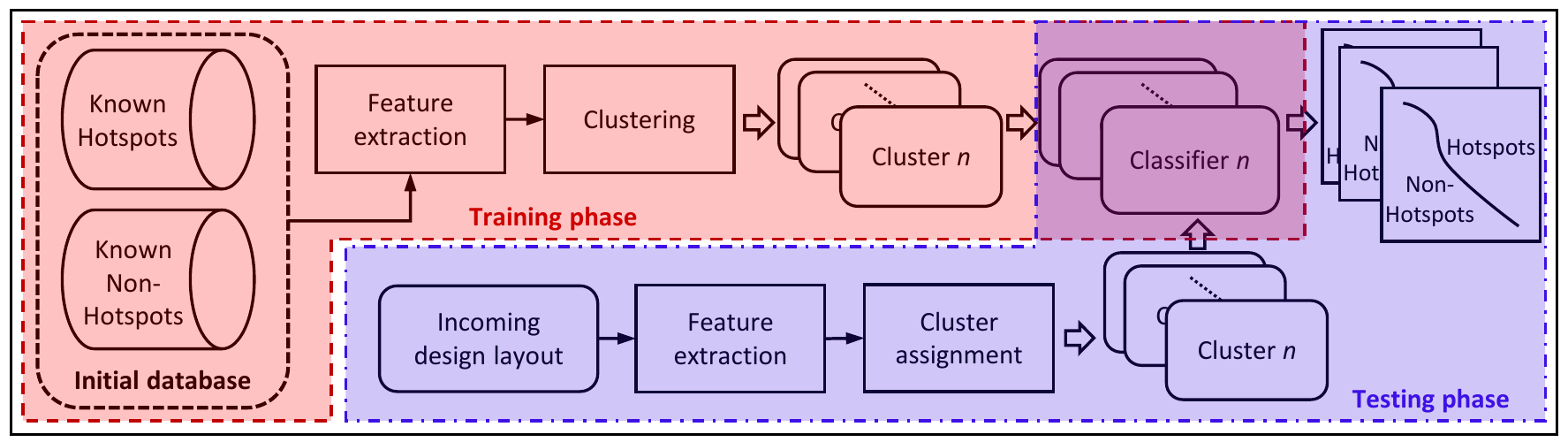}
	\caption{Previously proposed hotspot detection flow from \cite{MADKOUR_16}}
	\label{fig:isqed_flow}
\end{figure*}

\begin{table*}[]
\centering
\caption{Test results from ISQED'16 implementation (ETC patterns)}
\begin{tabular}{|l|c|c|c|c|c|c|c|c|c|c|c|c|}
\hline
\multicolumn{1}{|c|}{\multirow{2}{*}{\textbf{\begin{tabular}[c]{@{}c@{}}Test\\ Layout\end{tabular}}}} & \multicolumn{6}{c|}{\textbf{ISQED'16 (Non-enhanced)}} & \multicolumn{6}{c|}{\textbf{ISQED'16 (Enhanced)}} \\ \cline{2-13} 
\multicolumn{1}{|c|}{} & \textbf{\begin{tabular}[c]{@{}c@{}}HT hit\\ rate (\%)\end{tabular}} & \textbf{\begin{tabular}[c]{@{}c@{}}NHT hit\\ rate (\%)\end{tabular}} & \textbf{\begin{tabular}[c]{@{}c@{}}FP rate\\ (\%)\end{tabular}} & \textbf{\begin{tabular}[c]{@{}c@{}}FN rate\\ (\%)\end{tabular}} & \textbf{\begin{tabular}[c]{@{}c@{}}Total Err.\\ rate (\%)\end{tabular}} & \textbf{MCC} & \textbf{\begin{tabular}[c]{@{}c@{}}HT hit\\ rate (\%)\end{tabular}} & \textbf{\begin{tabular}[c]{@{}c@{}}NHT hit\\ rate (\%)\end{tabular}} & \textbf{\begin{tabular}[c]{@{}c@{}}FP rate\\ (\%)\end{tabular}} & \textbf{\begin{tabular}[c]{@{}c@{}}FN rate\\ (\%)\end{tabular}} & \textbf{\begin{tabular}[c]{@{}c@{}}Total Err.\\ rate (\%)\end{tabular}} & \textbf{MCC} \\ \hline
SPI & 91.45 & 98.19 & 1.77 & 0.15 & 1.92 & 0.65 & 91.73 & 99.08 & 0.90 & 0.15 & 1.05 & 0.76 \\ \hline
TV80 & 86.04 & 97.52 & 2.42 & 0.30 & 2.72 & 0.60 & 88.01 & 98.01 & 1.95 & 0.25 & 2.20 & 0.65 \\ \hline
AES & 86.69 & 97.67 & 2.28 & 0.32 & 2.60 & 0.63 & 88.17 & 97.95 & 2.00 & 0.28 & 2.28 & 0.66 \\ \hline
\textbf{Average} & \textbf{88.06} & \textbf{97.79} & \textbf{2.16} & \textbf{0.26} & \textbf{2.41} & \textbf{0.63} & \textbf{89.30} & \textbf{98.35} & \textbf{1.62} & \textbf{0.23} & \textbf{1.84} & \textbf{0.69} \\ \hline
\end{tabular}
\label{tab:res_isqed_ETC}
\end{table*}

\begin{table*}[]
\centering
\caption{Test results from ISQED'16 implementation (HTC patterns)}
\hspace*{-0.3cm}\begin{tabular}{|c|c|c|c|c|c|c|c|c|c|c|c|c|}
\hline
\multirow{2}{*}{\textbf{\begin{tabular}[c]{@{}c@{}}Test\\ Layout\end{tabular}}} & \multicolumn{6}{c|}{\textbf{ISQED'16 (Non-enhanced)}} & \multicolumn{6}{c|}{\textbf{ISQED'16 (Enhanced)}} \\ \cline{2-13} 
 & \textbf{\begin{tabular}[c]{@{}c@{}}HT hit\\ rate (\%)\end{tabular}} & \textbf{\begin{tabular}[c]{@{}c@{}}NHT hit\\ rate (\%)\end{tabular}} & \textbf{\begin{tabular}[c]{@{}c@{}}FP rate\\ (\%)\end{tabular}} & \textbf{\begin{tabular}[c]{@{}c@{}}FN rate\\ (\%)\end{tabular}} & \textbf{\begin{tabular}[c]{@{}c@{}}Tot. Err.\\ rate (\%)\end{tabular}} & \textbf{MCC} & \textbf{\begin{tabular}[c]{@{}c@{}}HT hit\\ rate (\%)\end{tabular}} & \textbf{\begin{tabular}[c]{@{}c@{}}NHT hit\\ rate (\%)\end{tabular}} & \textbf{\begin{tabular}[c]{@{}c@{}}FP rate\\ (\%)\end{tabular}} & \textbf{\begin{tabular}[c]{@{}c@{}}FN rate\\ (\%)\end{tabular}} & \textbf{\begin{tabular}[c]{@{}c@{}}Tot. Err.\\ rate (\%)\end{tabular}} & \textbf{MCC} \\ \hline
\multicolumn{1}{|l|}{HTC Patterns} & 97.57 & 19.76 & 40.34 & 1.21 & 41.54 & 0.28 & 99.30 & 75.05 & 12.54 & 0.35 & 12.89 & 0.77 \\ \hline
\end{tabular}
\label{tab:res_isqed_HTC}
\end{table*}

We also performed a study regarding the number of synthetic patterns necessary for DB enhancement. To aid this analysis, we used a non-enhanced dataset of 10,000 patterns and generated multiple enhanced datasets by varying the number of synthetic patterns generated for each hotspot pattern in the non-enhanced dataset from 0 to 520. The trained models were tested using a common testing dataset containing only HTC patterns. As shown in Figure \ref{fig:error_var}, increasing the number of synthetic patterns {\color{black} also increases the information-theoretic content of the training dataset and continues to reduce error. However, eventually we reach a point of diminishing returns. Plots, such as the one provided in Figure \ref{fig:error_var}, can inform the end-user who is seeking to make a trade-off between error reduction and lithography simulation overhead. We also note that performing lithography simulations is not necessarily a show-stopper, considering that it is a highly parallelizable one-time procedure.}
%increases the information-theoretic content in the training dataset and continues to reduce error. However, eventually, the rate of change of error tends to decrease. Therefore, the user needs to make a trade-off between reduction in error and the acceptable lithography simulation overhead. We note that, lithography simulation on synthetic patterns is a highly parallelizable one-time procedure.}
In Figure \ref{fig:error_var}, if we consider the case when no synthetic patterns are added as the baseline error, we observe that an addition of a mere 40 synthetic patterns (per known hotspot) reduces classification error by about 29.5\% (i.e., \% change from 42.75\% to 30.14\%). This result corroborates the significance, effectiveness and  practicality of the proposed synthetic database enhancement method.

%{\color{black}
%Furthermore, as expected, we observe that a constant increase the amount of synthetic patterns increases the information-theoretic content in the training dataset and continues to reduce error. However, the rate of change of error tends to decrease. Therefore, the user can make a trade-off between reduction in error and the acceptable lithography simulation overhead. We note that, lithography simulation on synthetic patterns is a highly parallelizable one-time procedure.
%}
%Gaurav- rephrased this statement: When we consider as the baseline error when no synthetic patterns are added, we can observe that an addition of a mere 40 synthetic patterns (per known hotspot) reduces classification error by about 29.48\% (\% change from 42.74\% to 30.14\%).

\subsection{Effectiveness of Database Enhancement on Previously Proposed Hotspot Detection Methods}\label{subsec:isqed_analysis}

In Section \ref{sec:intro}, we claimed that even previously proposed hotspot detection methods can benefit from the proposed database enhancement approach, therefore making it `method agnostic'. To demonstrate this quality of the proposed method, we implemented and experimented with a hotspot detection flow previously proposed in \cite{MADKOUR_16}, and a hotspot detection flow based on deep learning.

\begin{table*}[]
{\color{black}
\centering
\caption{Test results from a CNN-based implementation (ETC patterns)}
\begin{tabular}{|l|c|c|c|c|c|c|c|c|c|c|c|c|}

\hline
\multicolumn{1}{|c|}{\multirow{2}{*}{\textbf{\begin{tabular}[c]{@{}c@{}}Test\\ Layout\end{tabular}}}} & \multicolumn{6}{c|}{\textbf{CNN-based flow (Non-enhanced)}} & \multicolumn{6}{c|}{\textbf{CNN-based flow (Enhanced)}} \\ \cline{2-13} 
\multicolumn{1}{|c|}{} & \textbf{\begin{tabular}[c]{@{}c@{}}HT hit\\ rate (\%)\end{tabular}} & \textbf{\begin{tabular}[c]{@{}c@{}}NHT hit\\ rate (\%)\end{tabular}} & \textbf{\begin{tabular}[c]{@{}c@{}}FP rate\\ (\%)\end{tabular}} & \textbf{\begin{tabular}[c]{@{}c@{}}FN rate\\ (\%)\end{tabular}} & \textbf{\begin{tabular}[c]{@{}c@{}}Total Err.\\ rate (\%)\end{tabular}} & \textbf{MCC} & \textbf{\begin{tabular}[c]{@{}c@{}}HT hit\\ rate (\%)\end{tabular}} & \textbf{\begin{tabular}[c]{@{}c@{}}NHT hit\\ rate (\%)\end{tabular}} & \textbf{\begin{tabular}[c]{@{}c@{}}FP rate\\ (\%)\end{tabular}} & \textbf{\begin{tabular}[c]{@{}c@{}}FN rate\\ (\%)\end{tabular}} & \textbf{\begin{tabular}[c]{@{}c@{}}Total Err.\\ rate (\%)\end{tabular}} & \textbf{MCC} \\ \hline
SPI & 92.46 & 99.02 & 0.96 & 0.13 & 1.09 & 0.76 & 90.14 & 99.62 & 0.37 & 0.17 & 0.55 & 0.85 \\ \hline
TV80 & 88.04 & 98.57 & 1.40 & 0.25 & 1.65 & 0.70 & 86.59 & 99.41 & 0.58 & 0.28 & 0.86 & 0.81 \\ \hline
AES & 88.69 & 98.46 & 1.50 & 0.27 & 1.77 & 0.71 & 86.89 & 99.30 & 0.68 & 0.31 & 0.99 & 0.80 \\ \hline
\textbf{Average} & \textbf{89.73} & \textbf{98.68} & \textbf{1.29} & \textbf{0.22} & \textbf{1.50} & \textbf{0.72} & \textbf{87.87} & \textbf{99.44} & \textbf{0.54} & \textbf{0.25} & \textbf{0.80} & \textbf{0.82} \\ \hline

\end{tabular}
\label{tab:res_CNN_ETC}
}
\end{table*}

\begin{table*}[]
{\color{black}
\centering
\caption{Test results from a CNN-based implementation (HTC patterns)}
\hspace*{-0.3cm}\begin{tabular}{|c|c|c|c|c|c|c|c|c|c|c|c|c|}
\hline
\multirow{2}{*}{\textbf{\begin{tabular}[c]{@{}c@{}}Test\\ Layout\end{tabular}}} & \multicolumn{6}{c|}{\textbf{CNN-based flow (Non-enhanced)}} & \multicolumn{6}{c|}{\textbf{CNN-based flow (Enhanced)}} \\ \cline{2-13} 
 & \textbf{\begin{tabular}[c]{@{}c@{}}HT hit\\ rate (\%)\end{tabular}} & \textbf{\begin{tabular}[c]{@{}c@{}}NHT hit\\ rate (\%)\end{tabular}} & \textbf{\begin{tabular}[c]{@{}c@{}}FP rate\\ (\%)\end{tabular}} & \textbf{\begin{tabular}[c]{@{}c@{}}FN rate\\ (\%)\end{tabular}} & \textbf{\begin{tabular}[c]{@{}c@{}}Tot. Err.\\ rate (\%)\end{tabular}} & \textbf{MCC} & \textbf{\begin{tabular}[c]{@{}c@{}}HT hit\\ rate (\%)\end{tabular}} & \textbf{\begin{tabular}[c]{@{}c@{}}NHT hit\\ rate (\%)\end{tabular}} & \textbf{\begin{tabular}[c]{@{}c@{}}FP rate\\ (\%)\end{tabular}} & \textbf{\begin{tabular}[c]{@{}c@{}}FN rate\\ (\%)\end{tabular}} & \textbf{\begin{tabular}[c]{@{}c@{}}Tot. Err.\\ rate (\%)\end{tabular}} & \textbf{MCC} \\ \hline
\multicolumn{1}{|l|}{HTC Patterns} & 99.09 & 39.05 & 30.64 & 0.45 & 31.09 & 0.48 & 98.84 & 90.84 & 4.61 & 0.58 & 5.18 & 0.90 \\ \hline
\end{tabular}
\label{tab:res_CNN_HTC}
}
\end{table*}
\subsubsection{ISQED'16 implementation}\label{subsubsec:isqed}
In \cite{MADKOUR_16}, the authors posit that training a single classifier using the entire training dataset becomes too complicated and would result in high training times and significant performance degradation. Therefore, they proposed a divide-and-conquer approach, wherein they used topological clustering to divide the training dataset into many clusters (i.e., smaller datasets), with a separate classifier being trained for every cluster. To replicate this work, we implemented the flow shown in Figure \ref{fig:isqed_flow}. In this flow, we used the $k$-means algorithm \cite{kmeans} with a $k$ value of 10 for topological clustering. In the first experiment, we used the non-enhanced dataset for training, while in the second one we used the enhanced training dataset. In both cases, we used the testing dataset shown in Table \ref{tab:TestLayouts} as the common testing dataset. Similar to \cite{MADKOUR_16}, we performed hyper-parameter tuning using the grid-search method (both for the enhanced and the non-enhanced flow). The results from the non-enhanced flow serve as a baseline while the results from the enhanced flow show the improvement obtained solely due to database enhancement. The results from these experiments are shown separately for ETC patterns and HTC patterns in Tables \ref{tab:res_isqed_ETC} and \ref{tab:res_isqed_HTC}, respectively. In the case of ETC patterns, we observe that both flows perform similarly well by showing hotspot hit rates of about 89\% and false positive rates of about 2\%. In the case of HTC patterns, however, the non-enhanced flow shows very high false positives. The enhanced flow, on the other hand, shows about 69\% reduction in false positives (i.e., \% change from 40.34\% to 12.54\%). Thereby, this analysis demonstrates that the proposed database enhancement method can be adopted by previously proposed hotspot detection methods through minimal changes in their implementation, in order to obtain a significant reduction in false alarms.

%Similar to the observations made in Section \ref{subsec:exp_analysis}, we find that database enhancement provides a significant reduction in the number of false alarms when classifying HTC patterns. This analysis demonstrates that the proposed database enhancement method can be adopted by any of the previously proposed hotspot detection methods through minimal changes in their implementation, in order to obtain a significant reduction in false alarms.

\begin{figure}[]
	\centering
	\includegraphics[clip=true, trim=0.0in 0.05in 0.0in 0.0in,width=\linewidth, scale=1.0]{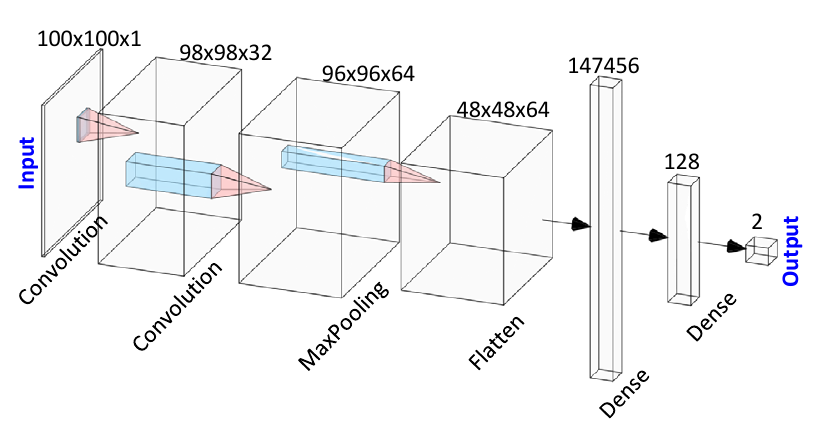}
	\vspace{-0.2in}
	\color{black}\caption{A CNN-Based implementation (adapted from \cite{MnistCNN})}
	\label{fig:cnn}
\end{figure}

{\color{black}
\subsubsection{CNN-based implementation}\label{subsubsec:cnn}
In the recent past, deep learning-based methods have been proposed for hotspot detection \cite{borisov2018lithography}\cite{yang2017DAC}\cite{yang2018TCAD}\cite{semisupervise2019}\cite{lithoroc2019}. To demonstrate the benefits of synthetic database enhancement in the realm of deep-learning, we implemented the CNN shown in Figure \ref{fig:cnn}. The network accepts a gray-scale image as its input. It consists of two stacked convolution layers, a 2D max-pooling layer, and a flattening layer which is followed by two densely connected layers. While the \textit{softmax} activation function is used in the final layer, \textit{relu} is used in the rest of the network. \textit{Dropout} is used for regularization. 

In this analysis, we used the same dataset shown in Tables \ref{tab:NonEnhTrainLayouts}, \ref{tab:EnhTrainLayouts}, and \ref{tab:TestLayouts}, but in the form of images. Patterns of size \SI{1000}{\nano\meter} $\times$ \SI{1000}{\nano\meter} were captured by centering on each of the layout fragments and, then, converted into images of size 100x100 pixels. While we used the non-enhanced training dataset in the first experiment, we used the enhanced dataset in the second. Both cases were tested using the common testing dataset shown in Table \ref{tab:TestLayouts}. Once again, the results from the non-enhanced flow serve as a baseline while the results from the enhanced flow show the improvement obtained solely due to synthetic database enhancement. The results from these experiments are shown separately for ETC patterns and HTC patterns in Tables \ref{tab:res_CNN_ETC} and \ref{tab:res_CNN_HTC}, respectively. The observations from these experiments are consistent with those from  previous sections. In case of ETC test patterns, we find that both enhanced and non-enhanced classifiers perform similarly well, achieving hotspot hit rates of about 89\% and false positive rates of about 1\%. In case of HTC patterns, however, we find that the non-enhanced classifier exhibits about 30\% false positives while the enhanced classifier reduces that amount to about 4.6\%. An 85\% reduction in false positives (i.e., \% change from 30.64\% to 4.61\%) demonstrates that even deep learning-based hotspot detection methods can significantly benefit from the proposed synthetic database enhancement.
}

 \begin{figure*}[t!]
	\vspace{-0.05in}
	\centering
	\pgfplotsset{width=0.9*\textwidth, height=6cm, compat=newest}
	\pgfplotsset{every axis/.append style={
			font=\footnotesize,
			line width=1.1pt,
			tick style={line width=0.6pt}}}
	%[width=0.5*\textwidth]
	%\hspace{-0.5in}
	\begin{tikzpicture}
	\begin{axis}[
	xlabel=Synthetic patterns per known hotspot,
	ylabel=Total Error \%,
	every axis y label/.style=
	{at={(ticklabel cs:0.5)},rotate=90,anchor=near ticklabel},
	nodes near coords,
	point meta=explicit symbolic,
	xtick=data,
	grid=major]
    \addplot+[smooth][color=blue, mark size=2pt, line width=1pt,
        mark=*,
        x=s,
        y=F,
        nodes near coords,
        point meta=explicit symbolic,
        visualization depends on={value \thisrow{anchor}\as\myanchor},,
        every node near coord/.append style={font=\footnotesize,anchor=\myanchor}
        %nodes={font=\small},
    %    nodes near coords align={anchor=west},
    ] table [row sep=\\,meta=Label]
    {
    s F Label anchor\\
		0 42.75   {42.75} east\\
		40 30.14  {30.14} east\\
		80 18.98  {18.98} 40\\
		120 15.78 {15.78} 50\\
		160 14.27 {14.27} 50\\
		200 13.17 {13.17} 50\\
		240 12.63 {12.63} 60\\
		280 12.03 {12.03} 60\\
		320 11.56 {11.56} 90\\
		360 11.23 {11.23} 60\\
		400 10.87 {10.87} 90\\
		440 10.62 {10.62} 60\\
		480 10.35 {10.35} 110\\
		520 10.17 {10.17} 60\\
    };
    
    \addplot+[smooth][color=red, mark size=2pt, line width=1pt,
        mark=*,
        x=s,
        y=F,
        nodes near coords,
        point meta=explicit symbolic,
        visualization depends on={value \thisrow{anchor}\as\myanchor},,
        every node near coord/.append style={font=\footnotesize,anchor=\myanchor}
        %nodes={font=\small},
    %    nodes near coords align={anchor=west},
    ] table [row sep=\\,meta=Label]
    {
    s F Label anchor\\
		0 42.75   {42.75} west\\
		40 36.94  {36.94} west\\
		80 21.88  {21.88} 220\\
		120 18.18 {18.18} 220\\
		160 16.2 {16.2} 220\\
		200 14.80 {14.80} 270\\
		240 13.86 {13.86} 240\\
		280 12.99 {12.99} 240\\
		320 12.37 {12.37} 240\\
		360 11.84 {11.84} 270\\
		400 11.31 {11.31} 260\\
		440 10.83 {10.83} 270\\
		480 10.56 {10.56} 250\\
		520 10.27 {10.27} 240\\
    };
    \legend{Random sampling, Active learning \cite{lin2018data}}
    \
	\end{axis}
	\end{tikzpicture}
	\color{black}\caption{Performance comparison of active learning against random sampling}
	\label{fig:error_var_active}
\end{figure*}
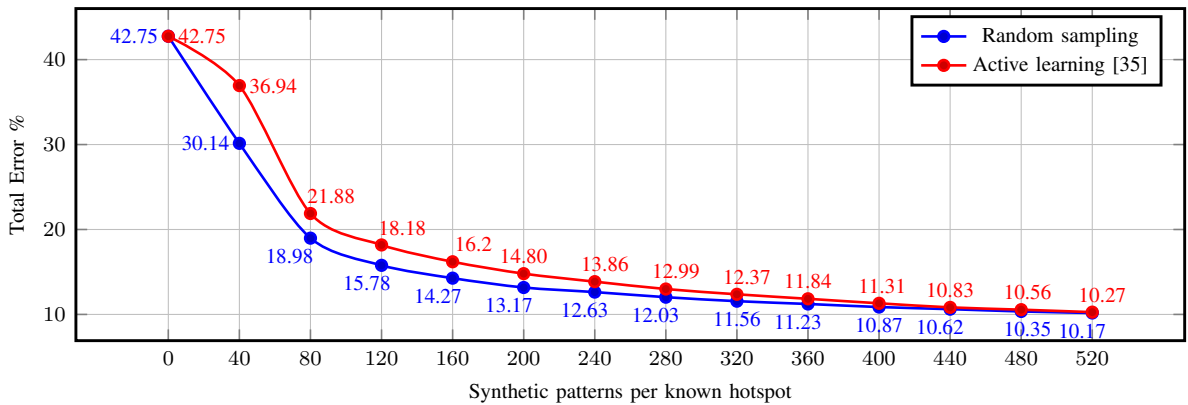

{\color{black}
\section{Discussion}\label{sec:discuss}

\subsection{Conciseness of Synthetic Dataset}\label{subsec:DiscussActiveLearn}
%Alternative title: Are all synthetic patterns essential?

Hotspot detection improvement through the proposed synthetic database enhancement method comes at the cost of computational effort for generating the synthetic patterns and performing lithography simulations. Therefore, it is important to ensure that the generated patterns are concise and non-redundant, in order to maximize the information-theoretic content that they contribute to the training dataset for this added cost. To evaluate such conciseness, we analyze the synthetically generated patterns using \textit{active learning}.    

%A robust synthetic pattern generation method must ensure that every pattern it generates is unique, and adds new information-theoretic content to the training dataset. To determine whether the proposed pattern generation method satisfies this property and to test whether there exists any redundancy in the generated synthetic dataset, we perform the following \textit{active learning-based} analysis.  

Active learning is a process which seeks to reduce the size of the training dataset without compromising the accuracy of the learned models. Such methods are particularly important when there is an abundance of unlabeled patterns but the process of labeling them is expensive and must, therefore, be used sparingly. In the context of hotspot detection, this reflects the cost of performing lithography simulations to label patterns as hotspots or non-hotspots. To this end, active learning flows employ advanced statistical methods \cite{semisupervise2019, lin2018data} which, at a fundamental level, sift through the unlabeled dataset and eliminate the redundant patterns (i.e., patterns which are similar to other patterns in the unlabeled/labeled datasets).%, thereby, reducing the number of patterns requiring lithography simulations.

Our conjecture is that, if there exists redundancy in the generated synthetic dataset, active learning should achieve the same performance (\% error) with fewer synthetic patterns than the complete dataset (or better performance with same number of synthetic patterns). To assess this conjecture, we implemented a recently-proposed active learning flow from \cite{lin2018data} and repeated the analysis shown in Figure \ref{fig:error_var}. Therein, in each step of the experiment and for every known hotspot we randomly sampled from a population of Independent and Identically Distributed (IID) synthetic variants. In contrast, in the repeated experiment, we use the synthetic patterns selected by the active learning method. The results of these two approaches are compared in Figure \ref{fig:error_var_active}.

As expected, at the extreme data-points both random sampling and active learning achieve the same error rate, since they produce the same training dataset. In all other cases, however, we observe that active learning results in higher error than random sampling. While this might seem counter-intuitive, such behavior is commonly observed when a statistical sampling method attempts to find similarities between data-points (patterns), when little to no such similarity exists between them \cite{westfall2013understanding}. In our case, the dissimilarity between synthetic variants mainly stems from the randomness in our synthetic pattern generation procedure which ensures that every new pattern has a different set/degree of variations in its features.

This analysis indicates that the patterns chosen by active learning were not necessarily more important than the rest, conclusively demonstrating that every sample generated using the proposed methodology is \textit{essential}, has new variation, and adds more information-theoretic content to the training dataset.

\subsection{Applicability to Newer Technology Nodes}\label{subsec:DiscussScaling}
%For our analysis, we relied on an open-source PDK (whose lithography models are publicly available), and open-source RTL codes to generate baseline design layouts. This was done mainly to facilitate independent result reproducibility and encourage further exploration of other ML-based hotspot detection methodologies which can leverage existing lithography models.

Owing to  their extremely complex fabrication processes, newer technology nodes --such as \SI{10}{\nano\meter} and \SI{7}{\nano\meter}-- introduce a large number of design constraints. Therefore, for our method to remain effective in synthetically enriching the information-theoretic content of  hotspot databases in these technologies, we must ensure that, despite these constraints, it continues to generate DRC-clean patterns\footnote{\color{black} Based on the results reported herein, we conjecture that, given sufficient DRC-clean synthetic variants of known hotspots, the ability of SOTA ML-based hotspot detection methods to learn the root cause is significantly improved in any technology. Regrettably, due to the lack of publicly available lithography models, we cannot apply and evaluate our entire flow in these newer technologies. For the same reason, we can also not evaluate our flow on the widely used ICCAD-2012 dataset \cite{CADcontest} or its recent derivative \cite{ReddyICCAD19}.}. To this end, we implemented and evaluated it using an industry-standard, Extreme UltraViolet Lithography (EUV)-based 7nm PDK \cite{ASAP7}.

%To verify whether the proposed method works with such constraints, we implement it using an industry-standard, Extreme UltraViolet Lithography (EUV)-based 7nm PDK \cite{ASAP7}. 

%Although we seek to implement and evaluate the entire flow shown in Figure \ref{fig:flow}, we are limited by the non-availability of lithography models for this PDK. Therefore, we perform the following limited analysis.

%In terms of scalability, as explained in the previous subsection, we have already demonstrated the effectiveness of the proposed methodology on the ICCAD-2019 benchmarks, which are from a 28nm technology node. However, we acknowledge that we observe much higher design constraints in very advanced technologies such as \SI{10}{\nano\meter} and \SI{7}{\nano\meter}. To verify whether the proposed methodology works with such constraints, we implement it using an industry-standard, Extreme UltraViolet Lithography (EUV)-based 7nm PDK \cite{ASAP7}. Although we seek to implement and evaluate the entire flow shown in Figure \ref{fig:flow}, we are limited by the non-availability of lithography models for this PDK. Therefore, we perform the following limited analysis.

Specifically, we first captured 1000 patterns from a full-chip design and used them as a proxy for the initial hotspot database. For every pattern in this dataset, we generated 200 synthetic variants using the methodology described in Section \ref{subsec:patgen}. We then subjected them to a full DRC test and found that approximately 41.97\% were DRC clean. 
This result demonstrates that the pattern variations (jogs, widths, spaces, etc.) which are carefully introduced by our method lead to legal patterns despite the more complex design constraints. While this percentage is lower than the 96\% DRC pass rate of synthetic patterns in \SI{45}{\nano\meter} technology, it is not a show-stopper. As explained in section \ref{subsec:exp_analysis}, synthetic pattern generation and the corresponding lithography simulations are highly parallelizable and, more importantly, one-time procedures. Therefore, with the understanding that slightly higher computational resources may be required due the increased complexity of the fabrication process, the proposed method remains highly applicable to newer technology nodes.

}% magenta color ending

\section{Conclusion}\label{sec:conclusion}
We discussed the problem of lithographic hotspots in advanced technology nodes, analyzed the state-of-the-art in this domain and highlighted their key limitation, namely the high false-alarm rate that they suffer from. To address this limitation, we proposed a novel database enhancement approach which involves synthetic pattern generation and design of experiments. We implemented the proposed flow using a 45nm PDK and experimentally demonstrated a reduction of up to 85\% in classification error, as compared to the state-of-the-art. Furthermore, we experimentally corroborated our conjecture that the proposed solution is method-agnostic and can be used by previously proposed ML-based hotspot detection methods in order to improve their performance.

\section*{Acknowledgments}\label{sec_akn}
This research has been partially supported by the Semiconductor Research Corporation (SRC) through task 2709.001.

\bibliographystyle{IEEE}
\bibliography{biblio}

% biography section
\begin{IEEEbiography}[{\includegraphics[width=1in,height=1.25in,clip,keepaspectratio]{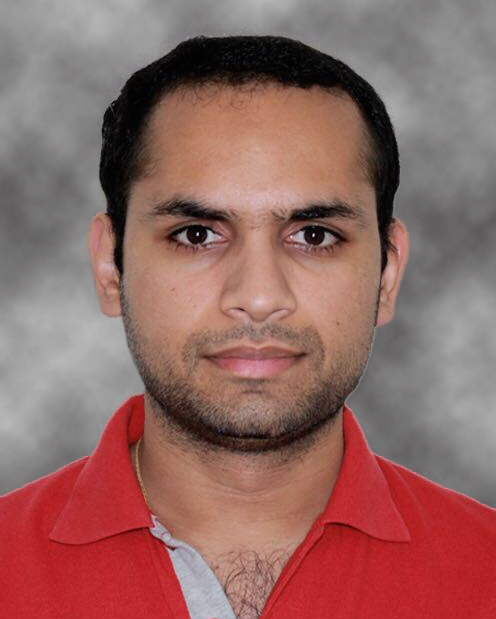}}]{Gaurav Rajavendra Reddy}
received a Bachelor of Engineering (BE) degree from the Visvesvaraya Technological University, India, in 2013. He received the MS and PhD degrees from the University of Texas at Dallas, USA, in 2019 and 2020, respectively. He worked as a post-silicon validation engineer at Tessolve, India, between 2013 and 2014. His research interests include applications of Machine Learning in Computer-Aided Design (CAD) and Design for Manufacturability (DFM).\end{IEEEbiography}

%\newpage 

\begin{IEEEbiography}[{\includegraphics[width=1in,height=1.25in,clip,keepaspectratio]{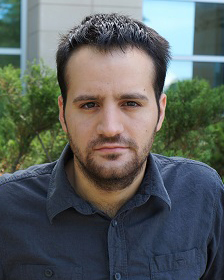}}]{Constantinos Xanthopoulos} received the B.S. degree in Computer Science from the University of Piraeus, Greece, in 2012, and the M.S. and Ph.D. degrees in Computer Engineering from The University of Texas at Dallas (UT Dallas),  in 2015 and 2019, respectively. 
%He is currently working toward a Ph.D. in Computer Engineering at UT Dallas, as a member of the Trusted and RELiable Architectures (TRELA) laboratory. 
His research interests focus on the application of statistical learning theory and machine learning to problems in analog test. He is a student member of the IEEE.\end{IEEEbiography}

\begin{IEEEbiography}[{\includegraphics[width=1in,height=1.25in,clip,keepaspectratio]{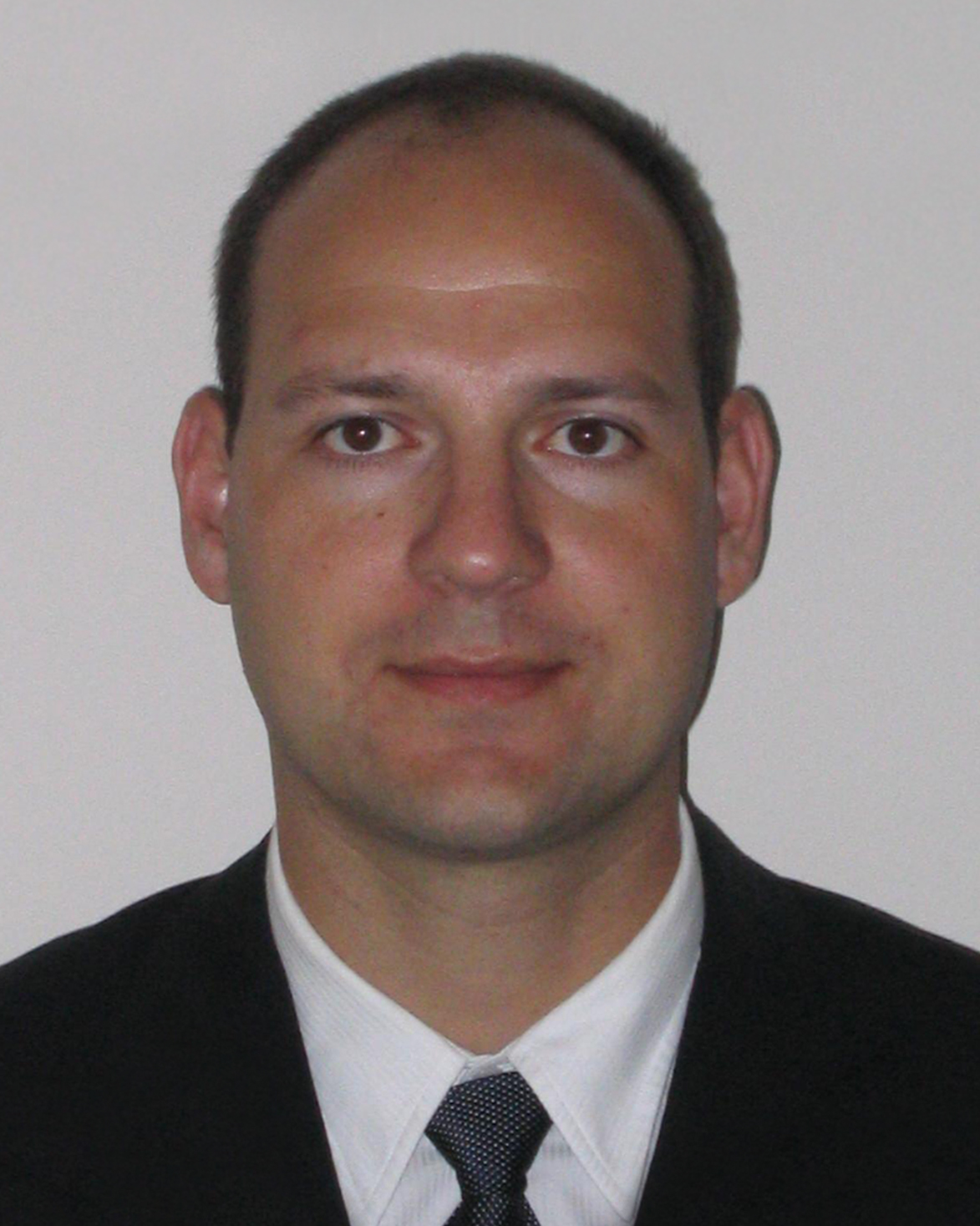}}]{Yiorgos Makris} (SM'08) received the Diploma of Computer Engineering from the University of Patras, Greece, in 1995 and the M.S. and Ph.D. degrees in Computer Engineering from the University of California, San Diego, in 1998 and 2001, respectively. After spending a decade on the faculty of Yale University, he joined UT Dallas where he is now a Professor of Electrical and Computer Engineering, leading the Trusted and RELiable Architectures (TRELA) Research Laboratory, and the Safety, Security and Healthcare thrust leader for Texas Analog Center of Excellence (TxACE). His research focuses on applications of machine learning and statistical analysis in the development of trusted and reliable integrated circuits and systems, with particular emphasis in the analog/RF domain. Prof. Makris serves as an Associate Editor of the IEEE Transactions on Computer-Aided Design of Integrated Circuits and Systems and has served as an Associate Editor for the IEEE Information Forensics and Security and the IEEE Design \& Test of Computers Periodical, and as a guest editor for the IEEE Transactions on Computers and the IEEE Transactions on Computer-Aided Design of Integrated Circuits and Systems. He is a recipient of the 2006 Sheffield Distinguished Teaching Award, Best Paper Awards from the 2013 IEEE/ACM Design Automation and Test in Europe (DATE'13) conference and the 2015 IEEE VLSI Test Symposium (VTS'15), as well as Best Hardware Demonstration Awards from the 2016 and the 2018 IEEE Hardware-Oriented Security and Trust Symposia (HOST'16 and HOST'18).
\end{IEEEbiography}

\end{document}